# A General Distributed Dual Coordinate Optimization Framework for Regularized Loss Minimization

**Shun Zheng*** ZHENGS14@MAILS.TSINGHUA.EDU.CN
*Institute for Interdisciplinary Information Sciences*
*Tsinghua University*
*Beijing, China*

**Jialei Wang** JIALEI@UCHICAGO.EDU
*Department of Computer Science*
*University of Chicago*
*Chicago, Illinois*

**Fen Xia** XIAFEN@BAIDU.COM
*Institute of Deep Learning*
*Baidu Research*
*Beijing, China*

**Wei Xu** WEIXU@TSINGHUA.EDU.CN
*Institute for Interdisciplinary Information Sciences*
*Tsinghua University*
*Beijing, China*

**Tong Zhang** TONGZHANG@TONGZHANG-ML.ORG
*Tencent AI Lab*
*Shenzhen, China*

## Abstract

In modern large-scale machine learning applications, the training data are often partitioned and stored on multiple machines. It is customary to employ the "data parallelism" approach, where the aggregated training loss is minimized without moving data across machines. In this paper, we introduce a novel distributed dual formulation for regularized loss minimization problems that can directly handle data parallelism in the distributed setting. This formulation allows us to systematically derive dual coordinate optimization procedures, which we refer to as *Distributed Alternating Dual Maximization* (DADM). The framework extends earlier studies described in (Boyd et al., 2011; Ma et al., 2017; Jaggi et al., 2014; Yang, 2013) and has rigorous theoretical analyses. Moreover with the help of the new formulation, we develop the accelerated version of DADM (Acc-DADM) by generalizing the acceleration technique from (Shalev-Shwartz and Zhang, 2014) to the distributed setting. We also provide theoretical results for the proposed accelerated version and the new result improves previous ones (Yang, 2013; Ma et al., 2017) whose iteration complexities grow linearly on the condition number. Our empirical studies validate our theory and show that our accelerated approach significantly improves the previous state-of-the-art distributed dual coordinate optimization algorithms.

*. Most of the work was done during the internship of Shun Zheng at Baidu Big Data Lab in Beijing.

## 1. Introduction

In large-scale machine learning applications for big data analysis, it becomes a common practice to partition the training data and store them on multiple machines that are connected via a commodity network. A typical setting of distributed machine learning is to allow these machines to train in parallel, with each machine processing its own local data with no data communication. This is often referred to as *data parallelism*. In order to reduce the overall training time, it is often necessary to increase the number of machines and to minimize the communication overhead. A major challenge is to reduce the training time as much as possible when we increase the number of machines. A practical solution requires two research directions: one is to improve the underlying system design making it suitable for machine learning algorithms (Dean and Ghemawat, 2008; Zaharia et al., 2012; Dean et al., 2012; Li et al., 2014); the other is to adapt traditional single-machine optimization methods to handle data parallelism (Boyd et al., 2011; Yang, 2013; Mahajan et al., 2013; Shamir et al., 2014; Jaggi et al., 2014; Mahajan et al., 2014; Ma et al., 2017; Takáč et al., 2015; Zhang and Lin, 2015). This paper focuses on the latter.

For big data machine learning on a single machine, there are generally two types of algorithms: batch algorithms such as gradient descent or L-BFGS (Liu and Nocedal, 1989), and stochastic optimization algorithms such as stochastic gradient descent and their modern variance reduced versions (Defazio et al., 2014; Johnson and Zhang, 2013). It is known that batch algorithms are relatively easy to parallelize. However, on a single machine, they converge more slowly than the modern stochastic optimization algorithms due to their high per-iteration computation costs. Specifically, it has been shown that the modern stochastic optimization algorithms converge faster than the traditional batch algorithms for convex regularized loss minimization problems. The faster convergence can be guaranteed in theory and observed in practice.

The fast convergence of modern stochastic optimization methods has led to studies to extend these methods to the distributed computing setting. Specifically, this paper considers the generalization of Stochastic Dual Coordinate Ascent method (Hsieh et al., 2008; Shalev-Shwartz and Zhang, 2013) and its proximal variant (Shalev-Shwartz and Zhang, 2014) to handle distributed training using data parallelism. Although this problem has been considered previously (Yang, 2013; Jaggi et al., 2014; Ma et al., 2017), these earlier approaches work with the dual formulation that is the same as the traditional single-machine dual formulation, where dual variables are coupled, and hence run into difficulties when they try to motivate and analyze the derived methods under the distributed environment.

A major contribution of this work is to introduce a new dual formulation specifically for distributed regularized loss minimization problems when data are distributed to multiple machines. In our new formulation, we decouple the local dual variables through introducing another dual variable $\beta$. This new dual formulation allows us to naturally extend the proximal SDCA algorithm (ProxSDCA) of (Shalev-Shwartz and Zhang, 2014) to the setting of multi-machine distributed optimization that can benefit from data parallelism. Moreover, the analysis of the original ProxSDCA can be easily adapted to the new formulation, leading to new theoretical results. This new dual formulation can also be combined with the acceleration technique of (Shalev-Shwartz and Zhang, 2014) to further improve convergence.

In the proposed formulation, each iteration of the distributed dual coordinate ascent optimization is naturally decomposed into a local step and a global step. In the local step, we allow the use of any local procedure to optimize a local dual objective function using local parameters and local data on each machine. This flexibility is similar to those of (Ma et al., 2017; Jaggi et al., 2014). For example, we may apply ProxSDCA as the local procedure. In the local step, a computer node can perform the optimization independently without communicating with each other. While in the global step, nodes communicate with each other to synchronize the local parameters and jointly update the global primal solution. Only this global step requires communication among nodes.

We summarize our main contributions as follows:

**New distributed dual formulation** This new formulation naturally leads to a two step local-global dual alternating optimization procedure for distributed machine learning. We thus call the resulting procedure *Distributed Alternating Dual Maximization* (DADM). Note that DADM directly generalizes ProxSDCA, which can handle complex regularizations such as $L_2$-$L_1$ regularization.

**New convergence analysis** The new formulation allows us to directly generalize the analysis of ProxSDCA in (Shalev-Shwartz and Zhang, 2014) to the distributed setting. This is in contrast to that of CoCoA$^+$ in (Ma et al., 2017), which employs a different analysis based on the $\Theta$-approximate solution assumption of the local solver. Our analysis can lead to simplified results in the commonly used mini-batch setup.

**Acceleration with theoretical guarantees** Based on the new distributed dual formulation, we can naturally derive a distributed version of the accelerated proximal SDCA method (AccProxSDCA) of (Shalev-Shwartz and Zhang, 2014), which has been shown to be effective on a single machine. We call the resulting procedure *Accelerated Distributed Alternating Dual Maximization* (Acc-DADM). The main idea is to modify the original formulation using a sequence of approximations that have stronger regularizations. Moreover We directly adapt theoretical analyses of AccProxSDCA to the distributed setting and provide guarantees for Acc-DADM. Our theorem guarantees that we can always obtain a computation speedup compared with the single-machine AccProxSDCA. This improves the theoretical results of DADM and previous methods (Yang, 2013; Ma et al., 2017) whose iteration complexities grow linearly on the condition number, and latter methods possibly fail to provide computation time improvement over the single-machine ProxSDCA when the condition number is large.

**Extensive empirical studies** We perform extensive experiments to compare the convergence and the scalability of the accelerated approach with that of previous state-of-the-art distributed dual coordinate ascent methods. Our empirical studies show that Acc-DADM can achieve faster convergence and better scalability than previous state-of-the-arts, in particular when the condition number is relatively large. This phenomenon is consistent with our theory.

We organize the rest of the paper as follows. Section 2 discusses related works. Section 3 provides preliminary definitions. Section 4 to 6 present the distributed primal formulation, the distributed dual formulation and our DADM method respectively. Section 7 then provides theorems for DADM. Section 8 introduces the accelerated version and provides corresponding theoretical guarantees. Section 9 includes all proofs of this paper. Section 10 provides extensive empirical studies of our novel method. Finally, Section 11 concludes the whole paper.

## 2. Related Work

Several generalizations of SDCA to the distributed settings have been proposed in the literature, including DisDCA (Yang, 2013), CoCoA (Jaggi et al., 2014), and CoCoA$^+$ (Ma et al., 2017).

DisDCA was the first attempt to study distributed SDCA, and it provided a basic theoretical analysis and a practical variant that behaves well empirically. Nevertheless, their theoretical result only applies to a few specially chosen mini-batch local dual updates that differ from the practical method used in their experiments. In particular, they did not show that optimizing each local dual problem leads to convergence. This limitation makes the methods they analyzed inflexible.

CoCoA was proposed to fix the above gap between theory and practice, and it was claimed to be a framework for distributed dual coordinate ascent in that it allows any local dual solver to be used for the local dual problem, rather than the impractical choices of DisDCA. However, the practical performance of CoCoA is inferior to the practical variant proposed in DisDCA with an aggressive local update. We note that the practical variant of DisDCA did not have a solid theoretical guarantee at that time.

CoCoA$^+$ fixed this situation and may be regarded as a generalization of CoCoA. The most effective choice of the aggregation parameter leads to a version which is similar to DisDCA, but allows exact optimization of each dual problem in their theory. According to studies in (Ma et al., 2017), the resulting CoCoA$^+$ algorithm performs significantly better than the original CoCoA both theoretically and empirically. The original CoCoA$^+$ (Ma et al., 2015) can only handle problems with the $L_2$ regularizer and it was generalized to general strongly convex regularizers in the long version (Ma et al., 2017). Besides (Smith et al., 2016) extended the framework to solve the primal problem of regularized loss minimization and cover general non-strongly convex regularizers such as $L_1$ regularizer, and (Hsieh et al., 2015) studied parallel SDCA with asynchronous updates.

Although CoCoA$^+$ has the advantage of allowing arbitrary local solvers and flexible approximate solutions of local dual problems, its theoretical analyses do not capture the contribution of the number of machines and the mini-batch size to the iteration complexity explicitly. Moreover the iteration complexities of both CoCoA$^+$ and DisDCA grow linearly with the condition number, thus they probably cannot provide computation time improvement over the single-machine SDCA when the condition number is large. This paper will remedy these unsatisfied aspects by providing a different analysis based on a new distributed dual formulation. Using this formulation, we can analyze procedures that can take an arbitrary local dual solver, which is like CoCoA$^+$; moreover, we allow the dual updates to be a mini-batch, which is like DisDCA. Moreover this formulation also allows us to naturally generalize AccProxSDCA and relevant theoretical results to the distributed setting. Our empirical results also validate the superiority of the accelerated approach.

While we focus on extending SDCA in this paper, we note that there are other approaches for parallel optimization. For example, there are direct attempts to parallelize stochastic gradient descent (Recht et al., 2011; Zinkevich et al., 2010). Some of these procedures only consider multi-core shared memory situation, which is very different from the distributed computing environment investigated in this paper. In the setting of distributed computing, data are partitioned into multiple machines and one often needs to study communication-efficient algorithms. In such cases, one extreme is to allow exact optimization of subproblems on each local machine as considered in (Shamir et al., 2014; Zhang and Lin, 2015). Although this approach minimizes communication, the computational cost for each local solver can dominate the overall training. Therefore in practice, it is necessary to do a trade-off by using the mini-batch update approach (Takac et al., 2013; Takáč et al., 2015). However, it is difficult for traditional mini-batch methods to design reasonable aggregation strategies to achieve fast convergence. (Takáč et al., 2015) studied how the step size can be reduced when the mini-batch size grows in the distributed setting. (Lee and Roth, 2015) derived an analytical solution of the optimal step size for dual linear support vector machine problems. Besides (Mahajan et al., 2013) presented a general framework for distributed optimization based on local functional approximation, which include several first-order and second-order methods as special cases, and (Mahajan et al., 2014) considered each machine to handle a block of coordinates, and proposed distributed block coordinate descent methods for solving $\ell_1$ regularized loss minimization problems.

Different from those methods, *Distributed Alternating Dual Maximization* (DADM) proposed in this work handles the trade-off between computation and communication by developing bounds for mini-batch dual updates, which is similar to (Yang, 2013). Moreover, DADM allows other better local solvers to achieve faster convergence in practice.

## 3. Preliminaries

In this section, we introduce some notations used in the following section. All functions that we consider in this paper are proper convex functions over a Euclidean space.

Given a function $f : \mathbb{R}^d \to \mathbb{R}$, we denote its **conjugate function** as

$$f^*(b) = \sup_a [b^\top a - f(a)].$$

A function $f : \mathbb{R}^d \to \mathbb{R}$ is **$L$-Lipschitz** with respect to $\|\cdot\|_2$ if for all $a, b \in \mathbb{R}^d$, we have

$$|f(a) - f(b)| \leq L\|a - b\|_2.$$

A function $f : \mathbb{R}^d \to \mathbb{R}$ is **$(1/\gamma)$-smooth** with respect to $\|\cdot\|_2$ if it is differentiable and its gradient is $(1/\gamma)$-Lipschitz with respect to $\|\cdot\|_2$. An equivalent definition is that for all $a, b \in \mathbb{R}^d$, we have

$$f(b) \leq f(a) + \nabla f(a)^\top (b - a) + \frac{1}{2\gamma}\|b - a\|_2^2.$$

A function $f : \mathbb{R}^d \to \mathbb{R}$ is **$\lambda$-strongly convex** with respect to $\|\cdot\|_2$ if for any $a, b \in \mathbb{R}^d$, we have

$$f(b) \geq f(a) + \nabla f(a)^\top (b - a) + \frac{\lambda}{2}\|b - a\|_2^2,$$

where $\nabla f(a)$ is any subgradient of $f(a)$.

It is well known that a function $f$ is $\gamma$-strongly convex with respect to $\|\cdot\|_2$ if and only if its conjugate function $f^*$ is $(1/\gamma)$-smooth with respect to $\|\cdot\|_2$.

## 4. Distributed Primal Formulation

In this paper, we consider the following generic regularized loss minimization problem:

$$\min_{w \in \mathbb{R}^d} \left[ P(w) := \sum_{i=1}^{n} \phi_i(X_i^\top w) + \lambda n g(w) + h(w) \right], \tag{1}$$

which is often encountered in practical machine learning problems. Here we assume each $X_i \in \mathbb{R}^{d \times q}$ is a $d \times q$ matrix, $w \in \mathbb{R}^d$ is the model parameter vector, $\phi_i(u)$ is a convex loss function defined on $\mathbb{R}^q$, which is associated with the $i$-th data point, $\lambda > 0$ is the regularization parameter, $g(w)$ is a strongly convex regularizer and $h(w)$ is another convex regularizer. A special case is to simply set $h(w) = 0$. Here we allow the more general formulation, which can be used to derive different distributed dual forms that may be useful for special purposes.

The above optimization formulation can be specialized to a variety of machine learning problems. As an example, we may consider the $L_2$-$L_1$ regularized least squares problem, where $\phi_i(x_i^\top w) = (w^\top x_i - y_i)^2$ for vector input data $x_i \in \mathbb{R}^d$ and real valued output $y_i \in \mathbb{R}$, $g(w) = \|w\|_2^2 + a\|w\|_1$, and $h(w) = b\|w\|_1$ for some $a, b \geq 0$.

If we set $h(w) = 0$, then it is well-known (see, for example, (Shalev-Shwartz and Zhang, 2014)) that the primal problem (1) has an equivalent single-machine dual form of

$$\max_{\alpha \in \mathbb{R}^n} \left[ D(\alpha) := -\sum_{i=1}^{n} \phi_i^*(-\alpha_i) - \lambda n g^* \left( \frac{\sum_{i=1}^n X_i \alpha_i}{\lambda n} \right) \right], \tag{2}$$

where $\alpha = [\alpha_1, \cdots, \alpha_n], \alpha_i \in \mathbb{R}^q$ $(i = 1, ..., n)$ are dual variables, $\phi_i^*$ is the convex conjugate function of $\phi_i$, and similarly, $g^*$ is the convex conjugate function of $g$.

The stochastic dual coordinate ascent method, referred to as SDCA in (Shalev-Shwartz and Zhang, 2014), maximizes the dual formulation (2) by optimizing one randomly chosen dual variable at each iteration. Throughout the algorithm, the following primal-dual relationship is maintained:

$$w(\alpha) = \nabla g^* \left( \frac{\sum_{i=1}^n X_i \alpha_i}{\lambda n} \right), \tag{3}$$

for some subgradient $\nabla g^*(v)$.

It is known that $w(\alpha^*) = w^*$, where $w^*$ and $\alpha^*$ are optimal solutions of the primal problem and the dual problem respectively. It was shown in (Shalev-Shwartz and Zhang, 2014) that the duality gap defined as $P(w(\alpha)) - D(\alpha)$, which is an upper-bound of the primal sub-optimality $P(w(\alpha)) - P(w^*)$, converges to zero. Moreover, a convergence rate can be established. In particular, for smooth loss functions, the convergence rate is linear.

We note that SDCA is suitable for optimization on a single machine due to the fact that it works with a dual formulation that is suitable for a single machine. In the following, we will generalize the single-machine dual formulation (2) to the distributed setting, and study the corresponding distributed version of SDCA.

In the distributed setting, we assume that the training data are partitioned and distributed to $m$ machines. In other words, the index set $S = \{1, ..., n\}$ of the training data is divided into $m$ non-overlapping partitions, where each machine $\ell \in \{1, ..., m\}$ contains its own partition $S_\ell \subseteq S$. We assume that $\cup_\ell S_\ell = S$, and we use $n_\ell := |S_\ell|$ to denote the size of the training data on machine $\ell$.

Next, we can rewrite the primal problem (1) as the following constrained minimization problem that is suitable for the multi-machine distributed setting:

$$\begin{aligned} &\min_{w;\{w_\ell\}_{\ell=1}^m} \sum_{\ell=1}^m P_\ell(w_\ell) + h(w) \\ &\quad s.t. \quad w_\ell = w, \quad \text{for all } \ \ell \in \{1, ..., m\}, \\ &\text{where} \quad P_\ell(w_\ell) := \sum_{i \in S_\ell} \phi_i(X_i^\top w_\ell) + \lambda n_\ell g(w_\ell), \end{aligned} \tag{4}$$

where $w_\ell$ represents the local primal variable on each machine $\ell$, $P_\ell$ is the corresponding local primal problem and the constraints $w_\ell = w$ are imposed to synchronize the local primal variables. Obviously this multi-machine distributed primal formulation (4) is equivalent to the original primal problem (1).

We note that the idea of objective splitting in (4) is similar to the global variable consensus formulation described in (Boyd et al., 2011). Instead of using the commonly used ADMM (Alternating Direction Method of Multipliers) method that is not a generalization of (2), in this paper we derive a distributed dual formulation based on (4) that directly generalizes (2). We further propose a framework called Distributed Alternating Dual Maximization (DADM) to solve the distributed dual formulation. One advantage of DADM over ADMM is that DADM does not need to solve the subproblems in high accuracy, and thus it can naturally enjoy the trade-off between computation and communication, which is similar to related methods such as DisDCA, CoCoA and CoCoA$^+$.

## 5. Distributed Dual Formulation

The optimization problem (4) can be further rewritten as:

$$\begin{aligned} &\min_{w;\{w_\ell\};\{u_i\}} \sum_{\ell=1}^m \left[ \sum_{i \in S_\ell} \phi_i(u_i) + \lambda n_\ell g(w_\ell) \right] + h(w) \\ &\quad s.t \quad u_i = X_i^\top w_\ell, \ \text{for all } i \in S_\ell \\ &\qquad\quad\ w_\ell = w, \ \text{for all } \ell \in \{1, ..., m\}. \end{aligned} \tag{5}$$

Here we introduce $n$ dual variables $\alpha := \{\alpha_i\}_{i=1}^n$, where each $\alpha_i$ is the Lagrange multiplier for the constraint $u_i - X_i^\top w_\ell = 0$, and $m$ dual variables $\beta := \{\beta_\ell\}_{\ell=1}^m$, where each $\beta_\ell$ is the Lagrange multiplier for the constraint $w_\ell - w = 0$. We can now introduce the primal-dual objective function with Lagrange multipliers as follows:

$$\begin{aligned} &J(w; \{w_\ell\}; \{u_i\}; \{\alpha_i\}; \{\beta_\ell\}) \\ :=& \sum_{\ell=1}^m \left[ \sum_{i \in S_\ell} \Big( \phi_i(u_i) + \alpha_i^\top (u_i - X_i^\top w_\ell) \Big) + \lambda n_\ell g(w_\ell) + \beta_\ell^\top (w_\ell - w) \right] + h(w). \end{aligned}$$

**Proposition 1** *Define the dual objective as*

$$D(\alpha,\beta) := \sum_{\ell=1}^{m} \left[ \sum_{i\in S_\ell} -\phi_i^*(-\alpha_i) - \lambda n_\ell g^* \left( \frac{\sum_{i\in S_\ell} X_i\alpha_i - \beta_\ell}{\lambda n_\ell} \right) \right] - h^* \left( \sum_\ell \beta_\ell \right).$$

*Then we have*

$$D(\alpha,\beta) = \min_{w;\{w_\ell\};\{u_i\}} J(w;\{w_\ell\};\{u_i\};\{\alpha_i\};\{\beta_\ell\}),$$

*where the minimizers are achieved when the following equations are satisfied*

$$\begin{aligned} \nabla\phi_i(u_i) + \alpha_i =& 0, \\ -\left( \sum_{i\in S_\ell} X_i\alpha_i - \beta_\ell \right) + \lambda n_\ell \nabla g(w_\ell) =& 0, \\ -\sum_\ell \beta_\ell + \nabla h(w) =& 0, \end{aligned} \tag{6}$$

*for some subgradients $\nabla\phi_i(u_i)$, $\nabla g(w_\ell)$, and $\nabla h(w)$.*

When $\beta = \{\beta_\ell\}$ are fixed, we may define the local single-machine dual formulation on each machine $\ell$ with respect to $\alpha_{(\ell)}$ as

$$\tilde{D}_\ell(\alpha_{(\ell)}|\beta_\ell) := \sum_{i\in S_\ell} -\phi_i^*(-\alpha_i) - \lambda n_\ell g^* \left( \frac{\sum_{i\in S_\ell} X_i\alpha_i - \beta_\ell}{\lambda n_\ell} \right), \tag{7}$$

where $\alpha_{(\ell)}$ represents local dual variables $\{\alpha_i; i\in S_\ell\}$ on machine $\ell$, $\beta_\ell \in \mathbb{R}^d$ serves as a carrier for synchronization of machine $\ell$. Based on Proposition 1, we obtain the following multi-machine distributed dual formulation for the corresponding primal problem (4):

$$D(\alpha,\beta) = \sum_{\ell=1}^{m} \tilde{D}_\ell(\alpha_{(\ell)}|\beta_\ell) - h^* \left( \sum_{\ell=1}^{m} \beta_\ell \right). \tag{8}$$

Moreover we have the non-negative duality gap, and zero duality gap can be achieved when $w$ is the minimizer of $P(w)$ and $(\alpha,\beta)$ maximizes the dual $D(\alpha,\beta)$.

**Proposition 2** *Given any $(w,\alpha,\beta)$, the following duality gap is non-negative:*

$$P(w) - D(\alpha,\beta) \geq 0.$$

*Moreover, zero duality gap can be achieved at $(w^*,\alpha^*,\beta^*)$, where $w^*$ is the minimizer of $P(w)$ and $(\alpha^*,\beta^*)$ is a maximizer of $D(\alpha,\beta)$.*

We note that the parameters $\{\beta_\ell\}_{\ell=1}^m$ pass the global information across multiple machines. When $\beta_\ell$ is fixed, $\tilde{D}_\ell(\alpha_{(\ell)}|\beta_\ell)$ with respect to $\alpha_{(\ell)}$ corresponds to the dual of the adjusted local primal problem:

$$\tilde{P}_\ell(w_\ell|\beta_\ell) := \sum_{i\in S_\ell} \phi_i(X_i^\top w_\ell) + \lambda n_\ell \tilde{g}_\ell(w_\ell), \tag{9}$$

where the original regularizer $\lambda n_\ell g(w_\ell)$ in $P_\ell(w_\ell)$ is replaced by the adjusted regularizer

$$\lambda n_\ell \tilde{g}_\ell(w_\ell) := \lambda n_\ell g(w_\ell) + \beta_\ell^\top w_\ell.$$

Similar to the single-machine primal-dual relationship of (3), we have the following local primal-dual relationship on each machine as:

$$w_\ell(\alpha_{(\ell)}, \beta_\ell) = \nabla g^*\left(\tilde{v}_\ell\right) = \nabla \tilde{g}_\ell^*\left(v_\ell\right), \tag{10}$$

where

$$v_\ell = \frac{\sum_{i \in S_\ell} X_i \alpha_i}{\lambda n_\ell}, \qquad \tilde{v}_\ell = v_\ell - \frac{\beta_\ell}{\lambda n_\ell}.$$

Moreover, we can define the global primal-dual relationship as

$$w(\alpha, \beta) = \nabla g^*\left(\tilde{v}\right) = \nabla \tilde{g}^*\left(v\right), \tag{11}$$

where

$$v = \frac{\sum_{i=1}^n X_i \alpha_i}{\lambda n}, \qquad \tilde{v} = v - \frac{\sum_\ell \beta_\ell}{\lambda n}.$$

We can also establish the relationship of global-local duality in Proposition 3.

**Proposition 3** *Given $(w, \alpha, \beta)$ and $\{w_\ell\}$ such that $w_1 = \cdots = w_m = w$, we have the following decomposition of global duality gap as the sum of local duality gaps:*

$$P(w) - D(\alpha, \beta) \geq \sum_{\ell=1}^m \left[\tilde{P}_\ell(w_\ell|\beta_\ell) - \tilde{D}_\ell(\alpha_{(\ell)}|\beta_\ell)\right],$$

*and the equality holds when $\nabla h(w) = \sum_\ell \beta_\ell$ for some subgradient $\nabla h(w)$.*

Although we allow arbitrary $h(w)$, the case of $h(w) = 0$ is of special interests. This corresponds to the conjugate function

$$h^*(\beta) = \begin{cases} +\infty & \text{if } \beta \neq 0 \\ 0 & \text{if } \beta = 0 \end{cases}.$$

That is, the term $h^*\left(\sum_{\ell=1}^m \beta_\ell\right)$ is equivalent to imposing the constraint $\sum_{\ell=1}^m \beta_\ell = 0$.

## 6. Distributed Alternating Dual Maximization

Minimizing the primal formulation (4) is equivalent to maximizing the dual formulation (8), and the latter can be achieved by repeatedly using the following alternating optimization strategy, which we refer to as **D**istributed **A**lternating **D**ual **M**aximization (**DADM**):

- **Local step**: fix $\beta_\ell$ and let each machine approximately optimize $\tilde{D}_\ell(\alpha_{(\ell)}|\beta_\ell)$ w.r.t $\alpha_{(\ell)}$ in parallel.
- **Global step**: maximize the global dual objective w.r.t $\beta_\ell$, and set the global primal parameter $w$ accordingly.

**Algorithm 1** Local Dual Update

Retrieve local parameters $(\alpha_\ell^{(t-1)}, \tilde{v}_\ell^{(t-1)})$
Randomly pick a mini-batch $Q_\ell \subset S_\ell$
Approximately maximize (12) w.r.t $\Delta\alpha_{Q_\ell}$
Update $\alpha_i^{(t)}$ as $\alpha_i^{(t)} = \alpha_i^{(t-1)} + \Delta\alpha_i$ for all $i \in Q_\ell$
**return** $\Delta v_\ell^{(t)} = \frac{1}{\lambda n_\ell} \sum_{i \in Q_\ell} X_i \Delta\alpha_i$

The above steps are applied in iterations $t = 1, 2, \ldots, T$. At the beginning of each iteration $t$, we assume that the local primal and dual variables on each local machine are $(\alpha_{(\ell)}^{(t-1)}, \beta_\ell^{(t-1)}, v_\ell^{(t-1)})$, then we seek to update $\alpha_{(\ell)}^{(t-1)}$ to $\alpha_{(\ell)}^{(t)}$ and $v_\ell^{(t-1)}$ to $v_\ell^{(t)}$ in the local step, and seek to update $\beta_\ell^{(t-1)}$ to $\beta_\ell^{(t)}$ in the global step.

We note that the local step can be executed in parallel w.r.t dual variables $\{\alpha_{(\ell)}\}_{\ell=1}^m$. In practice, it is often useful to optimize (7) approximately by using a randomly selected mini-batch $Q_\ell \subset S_\ell$ of size $|Q_\ell| = M_\ell$. That is, we want to find $\Delta\alpha_i^{(t)}$ with $i \in Q_\ell$ to approximately maximize the local dual objective as follows:

$$\tilde{D}_{Q_\ell}^{(t)}(\Delta\alpha_{Q_\ell}) := -\sum_{i \in Q_\ell} \phi_i^*(-\alpha_i^{(t-1)} - \Delta\alpha_i) - \lambda n_\ell g^* \left( \tilde{v}_\ell^{(t-1)} + \frac{\sum_{i \in Q_\ell} X_i \Delta\alpha_i}{\lambda n_\ell} \right). \tag{12}$$

This step is described in Algorithm 1. We can use any solver for this approximate optimization, and in our experiments, we choose ProxSDCA.

The global step is to synchronize all local solutions, which requires communication among the machines. This is achieved by optimizing the following dual objective with respect to all $\beta = \{\beta_\ell\}$:

$$\beta^{(t)} \in \arg\max_\beta D(\alpha^{(t)}, \beta). \tag{13}$$

**Proposition 4** *Given $v$, let $w(v)$ be the unique solution of the following optimization problem*

$$w(v) = \arg\min_w \left[ -\lambda n w^\top v + \lambda n g(w) + h(w) \right] \tag{14}$$

*that satisfies*

$$\lambda n \nabla g(w) + \nabla h(w) = \lambda n v$$

*for some subgradients $\nabla g(w)$ and $\nabla h(w) = \rho$ at $w = w(v)$. Then $\bar{\beta}(v) = \rho$ is a solution of*

$$\max_b \left[ -\lambda n g^* \left( v - \frac{b}{\lambda n} \right) - h^*(b) \right],$$

*and*

$$w(v) = \nabla g^* \left( v - \frac{\bar{\beta}(v)}{\lambda n} \right).$$

**Proposition 5** *Given $\alpha$, a solution of*

$$\max_{\beta} D(\alpha, \beta)$$

*can be obtained by setting*

$$\beta_\ell = \lambda n_\ell \left( v_\ell(\alpha_{(\ell)}) - v(\alpha) + \frac{\bar{\beta}(v(\alpha))}{\lambda n} \right)$$

*where $\bar{\beta}(v(\alpha))$ is defined in Proposition 4,*

$$v(\alpha) = \frac{\sum_{i=1}^{n} X_i \alpha_i}{\lambda n}, \qquad v_\ell(\alpha_{(\ell)}) = \frac{\sum_{i \in S_\ell} X_i \alpha_i}{\lambda n_\ell}.$$

*Moreover, if we let*

$$w = w(\alpha, \beta) = w(v(\alpha)) = \nabla g^* \left( v(\alpha) - \frac{\bar{\beta}(v(\alpha))}{\lambda n} \right),$$

*where $w(v)$ is defined in Proposition 4, and*

$$w_\ell = w_\ell(\alpha_{(\ell)}, \beta_\ell) = \nabla g^* \left( v_\ell(\alpha_{(\ell)}) - \frac{\beta_\ell}{\lambda n_\ell} \right),$$

*then $w = w_\ell$ for all $\ell$, and*

$$P(w) - D(\alpha, \beta) = \sum_{\ell=1}^{m} [\tilde{P}_\ell(w_\ell | \beta_\ell) - \tilde{D}_\ell(\alpha_{(\ell)} | \beta_\ell)].$$

According to Proposition 5, the solution of (13) is given by

$$\beta_\ell^{(t)} = \lambda n_\ell \left( v_\ell^{(t)} - v^{(t)} + \frac{\rho^{(t)}}{\lambda n} \right),$$

where

$$v^{(t)} = \sum_{\ell=1}^{m} \frac{n_\ell}{n} v_\ell^{(t)} = v^{(t-1)} + \sum_{\ell=1}^{m} \frac{n_\ell}{n} \Delta v_\ell^{(t)},$$

and $\rho^{(t)} = \nabla h(w^{(t)})$ is a subgradient of $h$ at the solution $w^{(t)}$ of

$$w^{(t)} = \arg\min_{w} \left[ -\lambda n w^\top v^{(t)} + \lambda n g(w) + h(w) \right],$$

that can achieve the first order optimality condition

$$-\lambda n v^{(t)} + \lambda n \nabla g(w^{(t)}) + \rho^{(t)} = 0$$

for some subgradient $\nabla g(w^{(t)})$.

**Algorithm 2** Distributed Alternating Dual Maximization (DADM)

**Input:** Objective $P(w)$, target duality gap $\epsilon$, warm start variables $w^{\text{init}}, \alpha^{\text{init}}, \beta^{\text{init}}, v^{\text{init}}$, (if not specified, set $w^{\text{init}} = 0, \alpha^{\text{init}} = 0, \beta^{\text{init}} = 0, v^{\text{init}} = 0$), .
**Initialize:** let $w^{(0)} = w^{\text{init}}$, $\alpha^{(0)} = \alpha^{\text{init}}$, $\beta^{(0)} = \beta^{\text{init}}$, $v^{(0)} = v^{\text{init}}$.
**for** $t = 1, 2, ...$ **do**
  (**Local step**)
  **for all machines** $\ell = 1, 2, ..., m$ **in parallel do**
    call an arbitrary local procedure, such as Algorithm 1
  **end for**
  (**Global step**)
  **Aggregate** $v^{(t)} = v^{(t-1)} + \sum_{\ell=1}^{m} \frac{n_\ell}{n} \Delta v_\ell^{(t)}$
  Compute $\tilde{v}^{(t)}$ according to (15)
  Let $\Delta\tilde{v}^{(t)} = \tilde{v}^{(t)} - \tilde{v}^{(t-1)}$
  **for all machines** $\ell = 1, 2, ..., m$ **in parallel do**
    update local parameter $\tilde{v}_\ell^{(t)} = \tilde{v}_\ell^{(t-1)} + \Delta\tilde{v}^{(t)}$
  **end for**
  **Stopping condition**: Stop if $P(w^{(t)}) - D(\alpha^{(t)}, \beta^{(t)}) \leq \epsilon$.
**end for**
**return** $w^{(t)} = \nabla g^*(\tilde{v}^{(t)})$, $\alpha^{(t)}$, $\beta^{(t)}$, $v^{(t)}$, and the duality gap $P(w^{(t)}) - D(\alpha^{(t)}, \beta^{(t)})$.

The definition of $\tilde{v}$ implies that after each global update, we have

$$\tilde{v}_\ell^{(t)} = \tilde{v}^{(t)} = v^{(t)} - \frac{\rho^{(t)}}{\lambda n} = \nabla g(w^{(t)}), \qquad \text{for all } \ell = 1, \ldots, m. \tag{15}$$

Since the objective (12) for the local step on each machine only depends on the mini-batch $Q_\ell$ (sampled from $S_\ell$) and the vector $\tilde{v}_\ell^{(t)}$, which needs to be synchronized at each global step, we know from (15) that at each time $t$, we can pass the same vector $\tilde{v}^{(t)}$ as $\tilde{v}_\ell^{(t)}$ to all nodes. In practice, it may be beneficial to pass $\Delta\tilde{v}^{(t)}$ instead, especially when $\Delta\tilde{v}^{(t)}$ is sparse but $\tilde{v}^{(t)}$ is dense. Put things together, the local-global DADM iterations can be summarized in Algorithm 2.

If we consider the special case of $h(w) = 0$, the solution of (15) is simply $\tilde{v}_\ell^{(t)} = \tilde{v}^{(t)} = v^{(t)}$, and the global step in Algorithm 2 can be simplified as first aggregating updates by

$$\Delta\tilde{v}^{(t)} = \Delta v^{(t)} = \sum_{\ell=1}^{m} \frac{n_\ell}{n} \Delta v_\ell^{(t)},$$

and then updating local parameters in parallel. Further, if $h(w) = 0$ and the data partition is balanced, that is $n_\ell$ are identical for all $\ell = 1, \ldots, m$, it can be verified that the DADM procedure (ignoring the mini-batch variation) is equivalent to CoCoA$^+$. Therefore the framework presented here may be regarded as an alternative interpretation.

Moreover, when the added regularization in (1) is complex and might involves more than one non-smooth term, considering the splitting of $g(w)$ and $h(w)$ can bring computational advantages. For example, to promote both sparsity and group sparsity in the predictor we often use the sparse group lasso regularization (Friedman et al., 2010), where a combination of $L_1$ norm and mixed $L_2/L_1$ norm (group sparse norm) is introduced: $\lambda_1 \sum_{\mathcal{G}} \|w_{\mathcal{G}}\|_2 + \lambda_2 \|w\|_1 + \lambda_3/2 \|w\|_2^2$, where we add a slight $L_2$ regularization to make it strongly convex, as did in (Shalev-Shwartz and Zhang, 2014). The proximal mapping with respect to the sparse group lasso regularization function does not have closed form solution, thus often relies on iterative minimization steps, but there are closed form proximal mapping with respect to either $L_2$-$L_1$ norm or the group norm. Thus if we simply set $h(w) = 0$ and $\lambda g(w) = \lambda_1 \sum_{\mathcal{G}} \|w_{\mathcal{G}}\|_2 + \lambda_2 \|w\|_1 + \lambda_3/2 \|w\|_2^2$, then both the local optimization update (12) and global synchronization step (14) will not have closed form solution. However, if we assign the group norm on $h(w)$ such that $h(w) = \lambda_1 \sum_{\mathcal{G}} \|w_{\mathcal{G}}\|_2$, and hence $\lambda g(w) = \lambda_2 \|w\|_1 + \lambda_3/2 \|w\|_2^2$, the local updates steps (12) will enjoy closed form update, which makes the implementation much easier and we only need to use iterative minimization on the (rare) global synchronization step (14).

## 7. Convergence Analysis

Let $w^*$ be the optimal solution for the primal problem $P(w)$ and $(\alpha^*, \beta^*)$ be the optimal solution for the dual problem $D(\alpha, \beta)$ respectively. For the primal solution $w^{(t)}$ and the dual solution $(\alpha^{(t)}, \beta^{(t)})$ at iteration $t$, we define the **primal sub-optimality** as

$$\epsilon_P^{(t)} := P(w^{(t)}) - P(w^*),$$

and the **dual sub-optimality** as

$$\epsilon_D^{(t)} := D(\alpha^*, \beta^*) - D(\alpha^{(t)}, \beta^{(t)}).$$

Due to the close relationship of the distributed dual formulation and the single-machine dual formulation, an analysis of DADM can be obtained by directly generalizing that of SDCA. We consider two kinds of loss functions, smooth loss functions that imply fast linear convergence and general $L$-Lipschitz loss functions. For the following two theorems we always assume that $g$ is 1-strongly convex w.r.t $\|\cdot\|_2$, $\|X_i\|_2^2 \leq R$ for all $i$, $M_\ell = |Q_\ell|$ is fixed on each machine, and our local procedure optimizes $\tilde{D}_{Q_\ell}^{(t)}$ sufficiently well on each machine such that $\tilde{D}_{Q_\ell}^{(t)}(\Delta\alpha_{Q_\ell}) \geq \tilde{D}_{Q_\ell}^{(t)}(\Delta\tilde{\alpha}_{Q_\ell})$, where $\Delta\tilde{\alpha}_{Q_\ell}$ is given by a special choice in each theorem.

**Theorem 6** *Assume that each $\phi_i$ is $(1/\gamma)$-smooth w.r.t $\|\cdot\|_2$ and $\Delta\tilde{\alpha}_{Q_\ell}$ is given by*

$$\Delta\tilde{\alpha}_i := s_\ell(u_i^{(t-1)} - \alpha_i^{(t-1)}), \qquad \text{for all } i \in Q_\ell,$$

*where $u_i^{(t-1)} := -\nabla\phi_i(X_i^\top w_\ell^{(t-1)})$ and $s_\ell := \frac{\gamma\lambda n_\ell}{\gamma\lambda n_\ell + M_\ell R} \in [0, 1]$. To reach an expected duality gap of $\mathbb{E}[P(w^{(T)}) - D(\alpha^{(T)}, \beta^{(T)})] \leq \epsilon$, every $T$ satisfying the following condition is sufficient,*

$$T \geq \left(\frac{R}{\gamma\lambda} + \max_\ell \frac{n_\ell}{M_\ell}\right) \log\left(\left(\frac{R}{\gamma\lambda} + \max_\ell \frac{n_\ell}{M_\ell}\right) \cdot \frac{\epsilon_D^{(0)}}{\epsilon}\right). \tag{16}$$

**Theorem 7** *Assume that each $\phi_i$ is L-Lipschitz w.r.t $\|\cdot\|_2$, and $\Delta\tilde{\alpha}_{Q_\ell}$ is given by*

$$\Delta\tilde{\alpha}_i := \frac{qn_\ell}{M_\ell}(u_i^{(t-1)} - \alpha_i^{(t-1)}), \quad \text{for all } i \in Q_\ell,$$

*where $-u_i^{(t-1)} := \nabla\phi_i(X_i^\top w_\ell^{(t-1)})$ and $q \in [0, \min_\ell(M_\ell/n_\ell)]$. To reach an expected normalized duality gap of $\mathbb{E}\left[\frac{P(\overline{w})-D(\overline{\alpha},\overline{\beta})}{n}\right] \leq \epsilon$, every $T$ satisfying the following condition is sufficient,*

$$T \geq T_0 + \max\left\{\tilde{n}, \frac{G}{\lambda\epsilon}\right\}, \tag{17}$$

$$T_0 \geq \max\left\{t_0, \frac{4G}{\lambda\epsilon} - 2\tilde{n} + t_0\right\}, \tag{18}$$

$$t_0 = \max\left\{0, \lceil \tilde{n}\log(2\lambda\tilde{n}\frac{\epsilon_D^{(0)}}{nG})\rceil\right\}, \tag{19}$$

*where $\tilde{n} = \max_\ell(n_\ell/M_\ell)$, $G = 4RL^2$ and $\overline{w}, \overline{\alpha}, \overline{\beta}$ represent either the average vector or a randomly chosen vector of $w^{(t-1)}, \alpha^{(t-1)}, \beta^{(t-1)}$ over $t \in \{T_0+1, ..., T\}$ respectively, such as $\overline{\alpha} = \frac{1}{T-T_0}\sum_{t=T_0+1}^{T}\alpha^{(t-1)}, \overline{\beta} = \frac{1}{T-T_0}\sum_{t=T_0+1}^{T}\beta^{(t-1)}, \overline{w} = \frac{1}{T-T_0}\sum_{t=T_0+1}^{T}w^{(t-1)}$.*

**Remark 8** *Both Theorem 6 and Theorem 7 incorporate two key components: the term $\max_\ell \frac{n_\ell}{M_\ell}$ and the condition number term $\frac{1}{\lambda\gamma}$ or $\frac{L^2}{\lambda}$. When the iteration complexity is dominated by the term $\max_\ell \frac{n_\ell}{M_\ell}$, we can speed up convergence and reduce the number of communications by increasing the number of machines $m$ or the local mini-batch size $M_\ell$. However, in some circumstances when the condition number is large, it will become the leading factor, and increasing $m$ or $M_\ell$ will not contribute to the computation speedup. To tackle this problem, we develop the accelerated version of DADM in Section 8.*

**Remark 9** *Our method is closely related to previous distributed extensions of SDCA. Our Theorem 6, 7 that provides theoretical guarantees for more general local updates achieves the same iteration complexity with the one in DisDCA that only allows some special choices of local mini-batch updates. Compared with the theorems of CoCoA$^+$ that are based on the $\Theta$-approximate solution of the local dual subproblem, although the derived bounds are within the same scale, $\tilde{O}(1/\epsilon)$ for Lipschitz losses and $\tilde{O}(\log(1/\epsilon))$ for smooth losses, our bounds are different and complementary. The analysis of CoCoA$^+$ can provide better insights for more accurate solutions of the local sub-problems. While our analysis is based on the mini-batch setup and can capture the contributions of the mini-batch size and the number of machines more explicitly.*

**Remark 10** *Since the bounds are derived with a special choice of $\Delta\tilde{\alpha}_{Q_\ell}$, the actual performance of the algorithm can be significantly better than what is indicated by the bounds when the local duals are better optimized. For example, we can choose ProxSDCA in (Shalev-Shwartz and Zhang, 2014) as the local procedure and adopt the sequential update strategy as the local solver of CoCoA$^+$ does. This is also the one used in our experiments.*

**Algorithm 3** Accelerated Distributed Alternating Dual Maximization (Acc-DADM).

**Prameters** $\kappa$, $\eta = \sqrt{\lambda/(\lambda + 2\kappa)}$, $\nu = (1-\eta)/(1+\eta)$.
**Initialize** $v^{(0)} = y^{(0)} = w^{(0)} = 0$, $\alpha^{(0)} = 0$, $\xi_0 = (1+\eta^{-2})(P(0) - D(0,0))$.
**for** $t = 1, 2, \ldots, T_{\text{outer}}$ **do**

1. **Construct new objective**:
$$P_t(w) = \sum_{i=1}^{n} \phi_i(X_i^\top w) + \lambda n g(w) + h(w) + \frac{\kappa n}{2} \left\| w - y^{(t-1)} \right\|_2^2.$$
2. **Call DADM solver**:
$$(w^{(t)}, \alpha^{(t)}, \beta^{(t)}, v^{(t)}, \epsilon_t) = \text{DADM}(P_t, (\eta \xi_{t-1})/(2 + 2\eta^{-2}), w^{(t-1)}, \alpha^{(t-1)}, \beta^{(t-1)}, v^{(t-1)}).$$
3. **Update**:
$$y^{(t)} = w^{(t)} + \nu(w^{(t)} - w^{(t-1)}).$$
4. **Update**:
$$\xi_t = (1 - \eta/2)\xi_{t-1}.$$

**end for**
Return $w^{(T_{\text{outer}})}$.

## 8. Acceleration

Theorem 6, 7 all imply that when the condition number $\frac{1}{\gamma\lambda}$ or $\frac{L^2}{\lambda}$ is relatively small, DADM converges fast. However, the convergence may be slow when the condition number is large and dominates the iteration complexity. In fact, we observe empirically that the basic DADM method converges slowly when the regularization parameter $\lambda$ is small. This phenomenon is also consistent with that of SDCA for the single-machine case. In this section, we introduce the Accelerated Distributed Alternating Dual Maximization (Acc-DADM) method that can alleviate the problem.

The procedure is motivated by (Shalev-Shwartz and Zhang, 2014), which employs an inner-outer iteration: at every iteration $t$, we solve a slightly modified objective, which adds a regularization term centered around the vector

$$y^{(t-1)} = w^{(t-1)} + \nu \left( w^{(t-1)} - w^{(t-2)} \right), \tag{20}$$

where $\nu \in [0, 1]$ is called the momentum parameter.

The accelerated DADM procedure (described in Algorithm 3) can be similarly viewed as an inner-outer algorithm, where DADM serves as the inner iteration, and in the outer iteration we adjust the regularization vector $y^{(t-1)}$. That is, at each outer iteration $t$, we define a modified local primal objective on each machine $\ell$, which has the same form as the original local primal objective (9), except that $\tilde{g}_\ell(w_\ell)$ is modified to $\tilde{g}_{\ell_t}(w_\ell)$ that is defined by

$$\begin{aligned} \lambda n_\ell \tilde{g}_{\ell_t}(w_\ell) =& \lambda n_\ell g_t(w_\ell) + \beta_\ell^\top w_\ell, \\ \lambda g_t(w_\ell) =& \lambda g(w_\ell) + \frac{\kappa}{2} \|w_\ell - y^{(t-1)}\|_2^2. \end{aligned}$$

It follows that we will need to solve a modified dual at each local step with $g^*(\cdot)$ replaced by $g_t^*(\cdot)$ in the local dual problem (12). Therefore, compared to the basic DADM procedure, nothing changes other than $g^*(\cdot)$ being replaced by $g_t^*(\cdot)$ at each iteration. Specifically, when the number of machines $m$ equals 1, this algorithm reduces to AccProxSDCA described in (Shalev-Shwartz and Zhang, 2014). Thus Acc-DADM can be naturally regarded as the distributed generalization of the single-machine AccProxSDCA. Moreover, Acc-DADM also allows arbitrary local procedures as DADM does.

Our empirical studies show that Acc-DADM significantly outperforms DADM in many cases. There are probably two reasons. One reason is the use of a modified regularizer $g_t(w)$ that is more strongly convex than the original regularizer $g(w)$ when $\kappa$ is much larger than $\lambda$. The other reason is closely related to the distributed setting considered in this paper. Observe that in the modified local primal objective

$$\tilde{P}_{\ell_t}(w_\ell|\beta_\ell) := \tilde{P}_\ell(w_\ell|\beta_\ell) + \frac{\kappa n_\ell}{2}\|w_\ell - y^{(t-1)}\|_2^2,$$

the first term corresponds to the original local primal objective and the second term is an extra regularization due to acceleration that constrains $w_\ell$ to be close to $y^{(t-1)}$. The effect is that different local problems become more similar to each other, which stabilize the overall system.

### 8.1 Theoretical Results of Acc-DADM for smooth losses

The following theorem establishes the computation efficiency guarantees for Acc-DADM.

**Theorem 11** *Assume that each $\phi_i$ is $(1/\gamma)$-smooth, and $g$ is 1-strongly convex w.r.t $\|\cdot\|_2$, $\|X_i\|_2^2 \leq R$ for all $i$, $M_\ell = |Q_\ell|$ is fixed on each machine. To obtain expected $\epsilon$ primal sub-optimality:*

$$\mathbb{E}[P(w^{(t)})] - P(w^*) \leq \epsilon,$$

*it is sufficient to have the following number of stages in Algorithm 3*

$$T_{\text{outer}} \geq 1 + \frac{2}{\eta}\log\left(\frac{\xi_0}{\epsilon}\right) = 1 + \sqrt{\frac{4(\lambda+2\kappa)}{\lambda}}\left(\log\left(\frac{2\lambda+2\kappa}{\lambda}\right) + \log\left(\frac{P(0)-D(0,0)}{\epsilon}\right)\right),$$

*and the number of inner iterations in DADM at each stage:*

$$T_{\text{inner}} \geq \left(\frac{R}{\gamma(\lambda+\kappa)} + \max_\ell \frac{n_\ell}{M_\ell}\right)\left(\log\left(\frac{R}{\gamma(\lambda+\kappa)} + \max_\ell \frac{n_\ell}{M_\ell}\right) + 7 + \frac{5}{2}\log\left(\frac{\lambda+2\kappa}{\lambda}\right)\right).$$

*In particular, suppose we assume $n_1 = n_2 = \ldots = n_m$, and $M_1 = M_2 = \ldots = M_m = b$, then the total vector computations for each machine is bounded by*

$$\tilde{O}(T_{\text{outer}}T_{\text{inner}}b) = \tilde{O}\left(\left(1+\sqrt{\frac{\kappa+\lambda}{\lambda}}\right)\left(\frac{R}{\gamma(\lambda+\kappa)} + \frac{n}{mb}\right)b\right).$$

**Remark 12** *When $\kappa = 0$, then the guarantees reduce to DADM. However, DADM only enjoys linear speedup over ProxSDCA when the number of machines satisfies $m \leq (n\gamma\lambda)/R$,* *and being able to obtain sub-linear speedup when $\frac{R}{\lambda\gamma} = \mathcal{O}(n)$. Besides enjoying the properties described above as DADM, if we choose $\kappa$ in Algorithm 3 as $\kappa = \frac{mR}{\gamma n} - \lambda$, and $b = 1$, then the total vector computations for each machine is bounded by*

$$\tilde{\mathcal{O}}\left(\sqrt{\frac{Rm}{\gamma n\lambda}}\left(\frac{n}{m}\right)\right) = \tilde{\mathcal{O}}\left(\sqrt{\frac{Rn}{\gamma\lambda m}}\right),$$

*which means Acc-DADM can be much faster than DADM when the condition number is large, and always obtain a square-root speedup over the single-machine AccProxSDCA.*

### 8.2 Acceleration for non-smooth, Lipschitz losses

Theorem 11 established rate of convergence for smooth loss functions, but the acceleration framework can be used on non-smooth, Lipschitz loss functions. The main idea is to use the Nesterov's smoothing technique (Nesterov, 2005) to construct a smooth approximation of the non-smooth function $\phi_i(\cdot)$, by adding a strongly-convex regularization term on the conjugate of $\phi_i(\cdot)$:

$$\tilde{\phi}_i^*(-\alpha_i) := \phi_i^*(-\alpha_i) + \frac{\gamma}{2}\left\|\alpha_i\right\|_2^2,$$

by the property of conjugate functions (e.g. Lemma 2 in (Shalev-Shwartz and Zhang, 2014)), we know $\tilde{\phi}_i(\cdot)$, as the conjugate function of $\tilde{\phi}_i^*(\cdot)$ is $(1/\gamma)$-smooth, and

$$0 \le \tilde{\phi}_i(u_i) - \phi_i(u_i) \le \frac{\gamma L^2}{2}.$$

Then instead of the original function with non-smooth losses (1), we minimize the smoothed objective:

$$\min_{w\in\mathbb{R}^d}\left[\hat{P}(w) := \sum_{i=1}^{n}\tilde{\phi}_i(X_i^\top w) + \lambda n g(w) + h(w)\right]. \tag{21}$$

The following corollary establishes the computation efficiency guarantees for Acc-DADM on non-smooth, Lipschitz loss functions.

**Corollary 13** *Assume that each $\phi_i$ is $L$-Lipschitz, and $g$ is 1-strongly convex w.r.t $\|\cdot\|_2$, $\|X_i\|_2^2 \le R$ for all $i$, $M_\ell = |Q_\ell|$ is fixed on each machine. To obtain expected $\epsilon$ normalized primal sub-optimality:*

$$\mathbb{E}\left[\frac{P(w^{(t)})}{n}\right] - \frac{P(w^*)}{n} \le \epsilon,$$

*it is sufficient to run Algorithm 3 on the smoothed objective (21), with*

$$\gamma = \frac{\epsilon}{L^2},$$

*and the following number of stages,*

$$T_{\text{outer}} \ge 1 + \frac{2}{\eta}\log\left(\frac{2\xi_0}{\epsilon}\right) = 1 + \sqrt{\frac{4(\lambda+2\kappa)}{\lambda}}\left(\log\left(\frac{2\lambda+2\kappa}{\lambda}\right) + \log\left(\frac{2(P(0)-D(0,0))}{\epsilon}\right)\right),$$

*and the number of inner iterations in DADM at each stage:*

$$T_{\text{inner}} \geq \left(\frac{L^2R}{\epsilon(\lambda+\kappa)} + \max_\ell \frac{n_\ell}{M_\ell}\right)\left(\log\left(\frac{L^2R}{\epsilon(\lambda+\kappa)} + \max_\ell \frac{n_\ell}{M_\ell}\right) + 7 + \frac{5}{2}\log\left(\frac{\lambda+2\kappa}{\lambda}\right)\right).$$

*In particular, suppose we assume $n_1 = n_2 = \ldots = n_m$, and $M_1 = M_2 = \ldots = M_m = b$, then the total vector computations for each machine is bounded by*

$$\tilde{\mathcal{O}}(T_{\text{outer}}T_{\text{inner}}b) = \tilde{\mathcal{O}}\left(\left(1+\sqrt{\frac{\kappa+\lambda}{\lambda}}\right)\left(\frac{L^2R}{\epsilon(\lambda+\kappa)} + \frac{n}{mb}\right)b\right).$$

**Remark 14** *When $\kappa = 0$, then the guarantees reduce to DADM for Lipschitz losses. Moreover, when $L^2Rm \geq n\epsilon\lambda$, if we choose $\kappa$ in Algorithm 3 as $\kappa = \frac{mL^2R}{n\epsilon} - \lambda$, and $b = 1$, then the total vector computation for each machine is bounded by*

$$\tilde{\mathcal{O}}\left(\sqrt{\frac{L^2Rm}{n\epsilon\lambda}}\left(\frac{n}{m}\right)\right) = \tilde{\mathcal{O}}\left(L\sqrt{\frac{Rn}{\epsilon\lambda m}}\right),$$

*which means Acc-DADM can be much faster than DADM when $\epsilon$ is small, and always obtain a square-root speedup over the single-machine AccProxSDCA.*

## 9. Proofs

In this section, we first present proofs about several previous propositions to establish our framework solidly. Then based on our new distributed dual formulation, we directly generalize the analysis of SDCA and adapt it to DADM in the commonly used mini-batch setup. Finally, we describe the proof for the theoretical guarantees of Acc-DADM.

### 9.1 Proof of Proposition 1

**Proof** Given any set of parameters $(w; \{w_\ell\}; \{u_i\}; \{\alpha_i\}; \{\beta_\ell\})$, we have

$$\begin{aligned}
&\min_{w;\{w_\ell\};\{u_i\}} J(w; \{w_\ell\}; \{u_i\}; \{\alpha_i\}; \{\beta_\ell\}) \\
&= \underbrace{\min_{w;\{w_\ell\}} \sum_{\ell=1}^m \left[\sum_{i\in S_\ell} \min_{u_i}\left(\phi_i(u_i) + \alpha_i^\top(u_i - X_i^\top w_\ell)\right) + \lambda n_\ell g(w_\ell) + \beta_\ell^\top(w_\ell - w)\right] + h(w)}_{A},
\end{aligned}$$

where the minimum is achieved at $\{u_i\}$ such that $\nabla\phi_i(u_i) + \alpha_i = 0$. By eliminating $u_i$ we obtain

$$\begin{aligned}
A &= \min_{w;\{w_\ell\}} \sum_{\ell=1}^m \left[\sum_{i\in S_\ell}\left(-\phi_i^*(-\alpha_i) - \alpha_i^\top X_i^\top w_\ell\right) + \lambda n_\ell g(w_\ell) + \beta_\ell^\top(w_\ell - w)\right] + h(w) \\
&= \underbrace{\min_w \sum_{\ell=1}^m \min_{w_\ell}\left[\sum_{i\in S_\ell} -\phi_i^*(-\alpha_i) - \left(\sum_{i\in S_\ell} X_i\alpha_i - \beta_\ell\right)^\top w_\ell + \lambda n_\ell g(w_\ell) - \beta_\ell^\top w\right] + h(w)}_{B},
\end{aligned}$$

where minimum is achieved at $\{w_\ell\}$ such that $-\left(\sum_{i\in S_\ell} X_i\alpha_i - \beta_\ell\right) + \lambda n_\ell \nabla g(w_\ell) = 0$. By eliminating $w_\ell$ we obtain

$$\begin{aligned} B =& \min_w \sum_{\ell=1}^m \left[\sum_{i\in S_\ell} -\phi_i^*(-\alpha_i) - \lambda n_\ell g^*\left(\frac{\sum_{i\in S_\ell} X_i\alpha_i - \beta_\ell}{\lambda n_\ell}\right) - \beta_\ell^\top w\right] + h(w) \\ =& \underbrace{\sum_{\ell=1}^m \left[\sum_{i\in S_\ell} -\phi_i^*(-\alpha_i) - \lambda n_\ell g^*\left(\frac{\sum_{i\in S_\ell} X_i\alpha_i - \beta_\ell}{\lambda n_\ell}\right)\right] - h^*\left(\sum_\ell \beta_\ell\right)}_{D(\alpha,\beta)}, \end{aligned}$$

where the minimizer is achieved at $w$ such that $-\sum_\ell \beta_\ell + \nabla h(w) = 0$. This completes the proof. ■

### 9.2 Proof of Proposition 2

**Proof** Given any $w$, if we take $u_i = X_i^\top w_\ell$ and $w_\ell = w$ for all $i$ and $\ell$, then $P(w) = J(w; \{w_\ell\}; \{u_i\}; \{\alpha_i\}; \{\beta_\ell\})$ for arbitrary $(\{\alpha_i\}; \{\beta_\ell\})$. It follows from Proposition 1 that

$$P(w) = J(w; \{w_\ell\}; \{u_i\}; \{\alpha_i\}; \{\beta_\ell\}) \geq D(\alpha, \beta).$$

$w^*$ is the minimizer of $P(w)$. When $w = w^*$, we may set $u_i = u_i^* = X_i^\top w^*$ and $w_\ell = w_\ell^* = w^*$. From the first order optimality condition, we can obtain

$$\sum_i X_i \nabla\phi_i(u_i^*) + \sum_\ell \lambda n_\ell \nabla g(w_\ell^*) + \nabla h(w^*) = 0.$$

If we take $\alpha_i^* = -\nabla\phi_i(u_i^*)$ and $\beta_\ell^* = \sum_{i\in S_\ell} X_i\alpha_i^* - \lambda n_\ell \nabla g(w_\ell^*)$ for some subgradients, then it is not difficult to check that all equations in (6) are satisfied. It follows that we can achieve equality in Proposition 1 as

$$P(w^*) = J(w^*; \{w_\ell^*\}; \{u_i^*\}; \{\alpha_i^*\}; \{\beta_\ell^*\}) = D(\alpha^*, \beta^*).$$

This means that zero duality gap can be achieved with $w^*$. It is easy to verify that $(\alpha^*, \beta^*)$ maximizes $D(\alpha, \beta)$, since for any $(\alpha, \beta)$, we have

$$\begin{aligned} D(\alpha, \beta) &\leq J(w^*; \{w_\ell^*\}; \{u_i^*\}; \{\alpha_i\}; \{\beta_\ell\}) \\ &= P(w^*) \\ &= D(\alpha^*, \beta^*). \end{aligned}$$

■

### 9.3 Proof of Proposition 3

**Proof** We have the decompositions

$$D(\alpha, \beta) = \sum_{\ell=1}^{m} \tilde{D}_\ell(\alpha_{(\ell)}|\beta_\ell) - h^*\left(\sum_\ell \beta_\ell\right),$$

and

$$P(w) = \sum_{\ell=1}^{m} \tilde{P}_\ell(w_\ell|\beta_\ell) - \left(\sum_\ell \beta_\ell\right)^\top w + h(w).$$

It follows that the duality gap

$$P(w) - D(\alpha, \beta) = \sum_{\ell=1}^{m} [\tilde{P}_\ell(w_\ell|\beta_\ell) - \tilde{D}_\ell(\alpha_{(\ell)}|\beta_\ell)] + h^*\left(\sum_\ell \beta_\ell\right) + h(w) - \left(\sum_\ell \beta_\ell\right)^\top w.$$

Note that the definition of convex conjugate function implies that

$$h^*\left(\sum_\ell \beta_\ell\right) + h(w) - \left(\sum_\ell \beta_\ell\right)^\top w \geq 0,$$

and the equality holds when $\nabla h(w) = \sum_\ell \beta_\ell$. This implies the desired result. ■

### 9.4 Proof of Proposition 4

**Proof** It is easy to check by using the duality that for any $b$ and $w$:

$$\begin{aligned} & -\lambda n g^*\left(v - \frac{b}{\lambda n}\right) - h^*(b) \\ \leq & \left[-\lambda n w^\top \left(v - \frac{b}{\lambda n}\right) + \lambda n g(w)\right] + \left[-b^\top w + h(w)\right] \\ = & -\lambda n w^\top v + \lambda n g(w) + h(w), \end{aligned}$$

and the equality holds if $b = \nabla h(w)$ and $v - \frac{b}{\lambda n} = \nabla g(w)$ for some subgradients. Based on the assumptions, the equality can be achieved at $b = \bar{\beta}(v) = \nabla h(w(v))$ and $w = w(v)$. This proves the desired result by noticing that $v - \frac{b}{\lambda n} = \nabla g(w)$ implies that $w = \nabla g^*(v - b/(\lambda n))$. ■

### 9.5 Proof of Proposition 5

**Proof** Since $\alpha$ is fixed, we know that the problem $\max_\beta D(\alpha, \beta)$ is equivalent to

$$\max_\beta \left[\sum_{\ell=1}^{m} -\lambda n_\ell g^*\left(v_\ell(\alpha_{(\ell)}) - \frac{\beta_\ell}{\lambda n_\ell}\right) - h^*\left(\sum_\ell \beta_\ell\right)\right].$$

Now by using Jensen's inequality, we obtain for any $(\beta'_\ell)$:

$$\begin{aligned}
&\sum_{\ell=1}^{m} -\lambda n_\ell g^*\left(v_\ell(\alpha_{(\ell)}) - \frac{\beta'_\ell}{\lambda n_\ell}\right) - h^*\left(\sum_\ell \beta'_\ell\right) \\
\leq& -\lambda n g^*\left(\sum_{\ell=1}^{m} \frac{n_\ell}{n} \frac{\sum_{i\in S_\ell} X_i\alpha_i - \beta'_\ell}{\lambda n_\ell}\right) - h^*\left(\sum_\ell \beta'_\ell\right) \\
=& -\lambda n g^*\left(v(\alpha) - \frac{\sum_\ell \beta'_\ell}{\lambda n}\right) - h^*\left(\sum_\ell \beta'_\ell\right) \\
\leq& -\lambda n g^*\left(v(\alpha) - \frac{\bar{\beta}(v(\alpha))}{\lambda n}\right) - h^*\left(\bar{\beta}(v(\alpha))\right).
\end{aligned} \tag{22}$$

In the above derivation, the last inequality has used Proposition 4. Here the equalities can be achieved when

$$v_\ell(\alpha_{(\ell)}) - \frac{\beta'_\ell}{\lambda n_\ell} = v(\alpha) - \frac{\bar{\beta}(v(\alpha))}{\lambda n}$$

for all $\ell$, which can be obtained with the choice of $\{\beta'_\ell\} = \{\beta_\ell\}$ given in the statement of the proposition. ■

### 9.6 Proof of Theorem 6

The following result is the mini-batch version of a related result in the analysis of ProxSDCA, which we apply to any local machine $\ell$. The proof is included for completeness.

**Lemma 15** *Assume that $\phi_i^*$ is $\gamma$-strongly convex w.r.t $\|\cdot\|_2$ (where $\gamma$ can be zero) and $g^*$ is 1-smooth w.r.t $\|\cdot\|_2$. Every local step, we randomly pick a mini-batch $Q_\ell \subset S_\ell$, whose size is $M_\ell := |Q_\ell|$, and optimize w.r.t dual variables $\alpha_i, i \in Q_\ell$. Then, using the simplified notation*

$$P_\ell(w_\ell^{(t-1)}) = \tilde{P}_\ell(w_\ell^{(t-1)}|\beta_\ell^{(t-1)}), \qquad D_\ell(\alpha_{(\ell)}^{(t-1)}) = \tilde{D}_\ell(\alpha_{(\ell)}^{(t-1)}|\beta_\ell^{(t-1)}),$$

*we have*

$$\mathbb{E}\left[D_\ell(\alpha_{(\ell)}^{(t)}) - D_\ell(\alpha_{(\ell)}^{(t-1)})\right] \geq \frac{s_\ell M_\ell}{n_\ell}\mathbb{E}\left[P_\ell(w_\ell^{(t-1)}) - D_\ell(\alpha_{(\ell)}^{(t-1)})\right] - \frac{s_\ell^2 M_\ell^2}{2\lambda n_\ell^2} G_\ell^{(t)}$$

*where*

$$\begin{aligned}
G_\ell^{(t)} &:= \sum_{i\in S_\ell}\left[\|X_i\|_2^2 - \frac{\gamma\lambda n_\ell(1-s_\ell)}{M_\ell s_\ell}\right]\mathbb{E}\left[\|u_i^{(t-1)} - \alpha_i^{(t-1)}\|_2^2\right] \\
\Delta\tilde{\alpha}_i &:= \alpha_i^{(t)} - \alpha_i^{(t-1)} = s_\ell(u_i^{(t-1)} - \alpha_i^{(t-1)}), \quad \textit{for all } i \in Q_\ell,
\end{aligned}$$

*and $-u_i^{(t-1)} = \nabla\phi_i(X_i^\top w_\ell^{(t-1)}), s_\ell \in [0,1]$.*

**Proof** Since only the elements in $Q_\ell$ are updated, the improvement in the dual objective can be written as

$$
\begin{aligned}
& D_\ell(\alpha_{(\ell)}^{(t)}) - D_\ell(\alpha_{(\ell)}^{(t-1)}) \\
= & \left( \sum_{i \in Q_\ell} -\phi_i^*(-\alpha_i^{(t)}) - \lambda n_\ell g^* \left( v_\ell^{(t-1)} + (\lambda n_\ell)^{-1} \sum_{i \in Q_\ell} X_i \Delta\tilde{\alpha}_i \right) \right) \\
& - \left( \sum_{i \in Q_\ell} -\phi_i^*(-\alpha_i^{(t-1)}) - \lambda n_\ell g^* \left( v_\ell^{(t-1)} \right) \right) \\
\geq & \underbrace{\left( \sum_{i \in Q_\ell} -\phi_i^*(-\alpha_i^{(t-1)} - \Delta\tilde{\alpha}_i) - \nabla g^*(v_\ell^{(t-1)})^\top \left( \sum_{i \in Q_\ell} X_i \Delta\tilde{\alpha}_i \right) - \frac{1}{2\lambda n_\ell} \| \sum_{i \in Q_\ell} X_i \Delta\tilde{\alpha}_i \|_2^2 \right)}_{A} \\
& - \underbrace{\left( \sum_{i \in Q_\ell} -\phi_i^*(-\alpha_i^{(t-1)}) \right)}_{B},
\end{aligned}
$$

where we have used the fact the $g^*$ is 1-smooth in the derivation of the inequality.

By the definition of the update in the algorithm, and the definition of $\Delta\tilde{\alpha}_i = s_\ell(u_i^{(t-1)} - \alpha_i^{(t-1)}), s_\ell \in [0, 1]$, we have

$$
\begin{aligned}
A \geq & \sum_{i \in Q_\ell} -\phi_i^*(-(\alpha_i^{(t-1)} + s_\ell(u_i^{(t-1)} - \alpha_i^{(t-1)})) \\
& - \nabla g^*(v^{(t-1)})^\top \left( \sum_{i \in Q_\ell} X_i s_\ell (u_i^{(t-1)} - \alpha_i^{(t-1)}) \right) \\
& - \frac{1}{2\lambda n_\ell} \| \sum_{i \in Q_\ell} X_i s_\ell (u_i^{(t-1)} - \alpha_i^{(t-1)}) \|_2^2
\end{aligned}
\tag{23}
$$

From now on, we omit the superscript $(t-1)$. Since $\phi_i^*$ is $\gamma$-strongly convex w.r.t $\| \cdot \|_2$, we have

$$
\begin{aligned}
\phi_i^*(-(\alpha_i + s_\ell(u_i - \alpha_i))) &= \phi^*(s_\ell(-u_i) + (1 - s_\ell)(-\alpha_i)) \\
&\leq s_\ell \phi^*(-u_i) + (1 - s_\ell)\phi_i^*(-\alpha_i) - \frac{\gamma}{2} s_\ell (1 - s_\ell) \|u_i - \alpha_i\|_2^2
\end{aligned}
\tag{24}
$$

Bringing Eq. (24) into Eq. (23), we get

$$
\begin{aligned}
A \geq & \sum_{i \in Q_\ell} \left( -s_\ell \phi_i^*(-u_i) - (1-s_\ell)\phi_i^*(-\alpha_i) + \frac{\gamma}{2} s_\ell (1-s_\ell) \|u_i - \alpha_i\|_2^2 \right) \\
& - w_\ell^\top \left( \sum_{i \in Q_\ell} s_\ell X_i (u_i - \alpha_i) \right) - \frac{1}{2\lambda n_\ell} \| \sum_{i \in Q_\ell} s_\ell X_i (u_i - \alpha_i) \|_2^2 \\
\geq & \underbrace{\left( \sum_{i \in Q_\ell} -\phi_i^*(-\alpha_i) \right)}_{B} + \sum_{i \in Q_\ell} \left( s_\ell \left( w_\ell^\top X_i(-u_i) - \phi_i^*(-u_i) \right) + s_\ell \phi_i^*(-\alpha_i) + s_\ell w_\ell^\top X_i \alpha_i \right) \\
& + \sum_{i \in Q_\ell} \frac{\gamma}{2} s_\ell (1-s_\ell) \|u_i - \alpha_i\|_2^2 - \sum_{i \in Q_\ell} \frac{M_\ell \|X_i(u_i - \alpha_i)\|_2^2}{2\lambda n_\ell} s_\ell^2,
\end{aligned}
$$

where we get the second inequality according to the fact that $\| \sum_{i \in Q_\ell} a_i \|_2^2 \leq \sum_{i \in Q_\ell} M_\ell \|a_i\|_2^2$.

Since we choose $-u_i = \nabla \phi_i(X_i^\top w_\ell)$, for some subgradients $\nabla \phi_i(X_i^\top w_\ell)$, which yields $w_\ell^\top X_i(-u_i) - \phi_i^*(-u_i) = \phi_i(X_i^\top w_\ell)$, then we obtain

$$
\begin{aligned}
A - B \geq & \sum_{i \in Q_\ell} s_\ell \left[ \phi_i(X_i^\top w_\ell) + \phi_i^*(-\alpha_i) + w_\ell^\top X_i \alpha_i \right] \\
& + \sum_{i \in Q_\ell} s_\ell \|u_i - \alpha_i\|_2^2 \left[ \frac{\gamma(1-s_\ell)}{2} - \frac{s_\ell M_\ell \|X_i\|_2^2}{2\lambda n_\ell} \right]. \\
= & \sum_{i \in Q_\ell} s_\ell \left[ \phi_i(X_i^\top w_\ell) + \phi_i^*(-\alpha_i) + w_\ell^\top X_i \alpha_i \right] \\
& + \frac{M_\ell}{2\lambda n_\ell} \sum_{i \in Q_\ell} s_\ell^2 \|u_i - \alpha_i\|_2^2 \left[ \frac{\gamma \lambda n_\ell (1-s_\ell)}{M_\ell s_\ell} - \|X_i\|_2^2 \right].
\end{aligned}
\tag{25}
$$

Recall that with $w_\ell = \nabla g^*(\tilde{v}_\ell)$, we have $g(w) + g^*(\tilde{v}) = w^\top \tilde{v}$. Then we derive the local duality gap as

$$
\begin{aligned}
& P_\ell(w_\ell) - D_\ell(\alpha_{(\ell)}) \\
= & \sum_{i \in S_\ell} \phi_i(X_i^\top w_\ell) + \lambda n_\ell g(w_\ell) + \beta_\ell^\top w_\ell - \left( \sum_{i \in S_\ell} -\phi_i^*(-\alpha_i) - \lambda n_\ell g^* \left( \frac{\sum_{i \in S_\ell} X_i \alpha_i - \beta_\ell}{\lambda n_\ell} \right) \right) \\
= & \sum_{i \in S_\ell} \left( \phi_i(X_i^\top w_\ell) + \phi_i^*(-\alpha_i) + w_\ell^\top X_i \alpha_i \right)
\end{aligned}
$$

Then, taking the expectation of Eq. (25) w.r.t the random choice of mini-batch set $Q_\ell$ at round $t$, we obtain

$$
\begin{aligned}
E_t[A_\ell - B_\ell] \geq & \frac{M_\ell}{n_\ell} \sum_{i \in S_\ell} s_\ell \left[\phi_i(X_i^\top w_\ell) + \phi_i^*(-\alpha_i) + w_\ell^\top X_i \alpha_i\right] \\
& + \frac{M_\ell^2}{2\lambda n_\ell^2} \sum_{i \in S_\ell} s_\ell^2 \|u_i - \alpha_i\|_2^2 \left[\frac{\gamma\lambda n_\ell (1-s_\ell)}{M_\ell s_\ell} - \|X_i\|_2^2\right] \\
= & \frac{s_\ell M_\ell}{n_\ell} \sum_{i \in S_\ell} \left[\phi_i(X_i^\top w_\ell) + \phi_i^*(-\alpha_i) + w_\ell^\top X_i \alpha_i\right] \\
& - \frac{M_\ell^2}{2\lambda n_\ell^2} \sum_{i \in S_\ell} s_\ell^2 \|u_i - \alpha_i\|_2^2 \left[\|X_i\|_2^2 - \frac{\gamma\lambda n_\ell (1-s_\ell)}{M_\ell s_\ell}\right].
\end{aligned}
$$

Take expectation of both sides w.r.t the randomness in previous iterations, we have

$$
\mathbb{E}[A_\ell - B_\ell] \geq \frac{s_\ell M_\ell}{n_\ell} \mathbb{E}\ \left[P_\ell(w_\ell) - D_\ell(\alpha_{(\ell)})\right] - \frac{s_\ell^2 M_\ell^2}{2\lambda n_\ell^2} G_\ell^{(t)},
$$

where

$$
G_\ell^{(t)} := \sum_{i \in S_\ell} \left[\|X_i\|_2^2 - \frac{\gamma\lambda n_\ell (1-s_\ell)}{M_\ell s_\ell}\right] \mathbb{E}\ \left[\|u_i - \alpha_i\|_2^2\right].
$$

∎

**Proof of Theorem 6.**
**Proof** We will apply Lemma 15 with

$$
s_\ell = \frac{1}{1 + \frac{RM_\ell}{\gamma\lambda n_\ell}} = \frac{\gamma\lambda n_\ell}{\gamma\lambda n_\ell + M_\ell R} \in [0,1], i \in S_\ell.
$$

Recall that $\|X_i\|_2^2 \leq R$ for all $i \in S_\ell$, then we have

$$
\|X_i\|_2^2 - \frac{\gamma\lambda n_\ell (1-s_\ell)}{M_\ell s_\ell} \leq 0, \text{ for all } i \in S_\ell,
$$

which implies that $G_\ell^{(t)} \leq 0$ for all $\ell$. It follows that for all $\ell$ after the local update step we have:

$$
\begin{aligned}
& \mathbb{E}\ [\tilde{D}_\ell(\alpha_{(\ell)}^{(t)} | \beta_\ell^{(t-1)}) - \tilde{D}_\ell(\alpha_{(\ell)}^{(t-1)} | \beta_\ell^{(t-1)})] \\
\geq & \frac{s_\ell M_\ell}{n_\ell} \mathbb{E}\ \left[\tilde{P}_\ell(w_\ell^{(t-1)} | \beta_\ell^{(t-1)}) - \tilde{D}_\ell(\alpha_{(\ell)}^{(t-1)} | \beta_\ell^{(t-1)})\right].
\end{aligned} \tag{26}
$$

Now we note that after the global step at iteration $t-1$, the choices of $w^{(t-1)}$ and $\beta^{(t-1)}$ in DADM is according to the choice of Proposition 4 and Proposition 5, it follows from Proposition 5 that the following relationship between the global and local duality gap at the beginning of the $t$-th iteration is satisfied:

$$P(w^{(t-1)}) - D(\alpha^{(t-1)}, \beta^{(t-1)}) = \sum_{\ell} \left[ \tilde{P}_\ell(w_\ell^{(t-1)} | \beta_\ell^{(t-1)}) - \tilde{D}_\ell(\alpha_{(\ell)}^{(t-1)} | \beta_\ell^{(t-1)}) \right].$$

Using this decomposition and summing over $\ell$ in (26), we obtain

$$\mathbb{E}\,[D(\alpha^{(t)}, \beta^{(t-1)}) - D(\alpha^{(t-1)}, \beta^{(t-1)})] \geq q\mathbb{E}\,[P(w^{(t-1)}) - D(\alpha^{(t-1)}, \beta^{(t-1)})],$$

where

$$q = \min_\ell \frac{s_\ell M_\ell}{n_\ell} = \min_\ell \frac{\gamma\lambda M_\ell}{\gamma\lambda n_\ell + M_\ell R}.$$

Since $D(\alpha^{(t)}, \beta^{(t)}) \geq D(\alpha^{(t)}, \beta^{(t-1)})$, we obtain

$$\mathbb{E}\,[D(\alpha^{(t)}, \beta^{(t)}) - D(\alpha^{(t-1)}, \beta^{(t-1)})] \geq q\mathbb{E}\,[P(w^{(t-1)}) - D(\alpha^{(t-1)}, \beta^{(t-1)})].$$

Let $(\alpha^*, \beta^*)$ be the optimal solution of the dual problem, we have defined the dual suboptimality as $\epsilon_D^{(t)} := D(\alpha^*, \beta^*) - D(\alpha^{(t)}, \beta^{(t)})$. Let $\epsilon_G^{(t-1)} = P(w^{(t-1)}) - D(\alpha^{(t-1)}, \beta^{(t-1)})$, and we know that $\epsilon_D^{(t-1)} \leq \epsilon_G^{(t-1)}$. It follows that

$$\mathbb{E}[\epsilon_D^{(t-1)}] \geq \mathbb{E}[\epsilon_D^{(t-1)} - \epsilon_D^{(t)}] \geq q\mathbb{E}[\epsilon_G^{(t-1)}] \geq q\mathbb{E}[\epsilon_D^{(t-1)}].$$

Therefore we have

$$q\mathbb{E}[\epsilon_G^{(t)}] \leq \mathbb{E}[\epsilon_D^{(t)}] \leq (1-q)\mathbb{E}[\epsilon_D^{(t-1)}] \leq (1-q)^t \epsilon_D^{(0)} \leq e^{-qt}\epsilon_D^{(0)}.$$

To obtain an expected duality gap of $E[\epsilon_G^{(T)}] \leq \epsilon$, every $T$, which satisfies

$$T \geq \frac{1}{q} \log\left(\frac{1}{q}\frac{\epsilon_D^{(0)}}{\epsilon}\right),$$

is sufficient. This proves the desired bound. ■

### 9.7 Proof of Theorem 7

Now, we consider $L$-Lipschitz loss functions and use the following basic lemma for $L$-Lipschitz losses taken from (Shalev-Shwartz and Zhang, 2013, 2014).

**Lemma 16** *Let $\phi : \mathbb{R}^q \to \mathbb{R}$ be an $L$-Lipschitz function w.r.t $\|\cdot\|_2$, then we have $\phi^*(\alpha) = \infty$, for any $\alpha \in \mathbb{R}^q$ s.t. $\|\alpha\|_2 > L$.*

**Proof of Theorem 7.**
**Proof** Applying Lemma 15 with $\gamma = 0$, then we have

$$G_\ell^{(t)} = \sum_{i \in S_\ell} \|X_i\|_2^2 \; \mathbb{E}\left[\|u_i^{(t-1)} - \alpha_i^{(t-1)}\|_2^2\right].$$

According to Lemma 16, we know that $\|u_i^{(t-1)}\|_2 \le L$ and $\|\alpha_i^{(t-1)}\|_2 \le L$, thus we have

$$\|u_i^{(t-1)} - \alpha_i^{(t-1)}\|_2^2 \le 2\left(\|u_i^{(t-1)}\|_2^2 + \|\alpha_i^{(t-1)}\|_2^2\right) \le 4L^2.$$

Recall that $\|X_i\|_2^2 \le R$, then we have $G_\ell^{(t)} \le G_\ell$, where $G_\ell = 4n_\ell RL^2$. Combining this into Lemma 15, we have

$$\begin{aligned}&\mathbb{E}\,[\tilde{D}_\ell(\alpha_{(\ell)}^{(t)}|\beta_\ell^{(t-1)}) - \tilde{D}_\ell(\alpha_{(\ell)}^{(t-1)}|\beta_\ell^{(t-1)})]\\ \ge&\frac{s_\ell M_\ell}{n_\ell}\mathbb{E}\,\left[\tilde{P}_\ell(w_\ell^{(t-1)}|\beta_\ell^{(t-1)}) - \tilde{D}_\ell(\alpha_{(\ell)}^{(t-1)}|\beta_\ell^{(t-1)})\right] - \frac{s_\ell^2 M_\ell^2}{2\lambda n_\ell^2}G_\ell. \end{aligned} \tag{27}$$

Now we also note that after the global step at iteration $t-1$, the choices of $w^{(t-1)}$ and $\beta^{(t-1)}$ in DADM is according to the choice of Proposition 4 and Proposition 5, it follows from Proposition 5 that the following relationship of global and local duality gap at the beginning of the $t$-th iteration is satisfied:

$$P(w^{(t-1)}) - D(\alpha^{(t-1)}, \beta^{(t-1)}) = \sum_\ell \left[\tilde{P}_\ell(w_\ell^{(t-1)}|\beta_\ell^{(t-1)}) - \tilde{D}_\ell(\alpha_{(\ell)}^{(t-1)}|\beta_\ell^{(t-1)})\right].$$

Summing the inequality (27) over $\ell$, combining with the above decomposition and bringing $D(\alpha^{(t)}, \beta^{(t)}) \ge D(\alpha^{(t)}, \beta^{(t-1)})$ into it, we get

$$\mathbb{E}[D(\alpha^{(t)}, \beta^{(t)}) - D(\alpha^{(t-1)}, \beta^{(t-1)})] \ge q\mathbb{E}[P(w^{(t-1)}) - D(\alpha^{(t-1)}, \beta^{(t-1)})] - \sum_{\ell=1}^m \frac{q^2}{2\lambda}G_\ell, \tag{28}$$

where $q \in [0, \min_\ell \frac{M_\ell}{n_\ell}], q = \frac{s_\ell M_\ell}{n_\ell}$ and $s_\ell \in [0,1]$ is chosen so that all $\frac{s_\ell M_\ell}{n_\ell}(\ell = 1, ..., m)$ are equal.

Let $(\alpha^*, \beta^*)$ be the optimal solution for the dual problem $D(\alpha, \beta)$, and we have defined the dual suboptimality as $\epsilon_D^{(t)} := D(\alpha^*, \beta^*) - D(\alpha^{(t)}, \beta^{(t)})$. Note that the duality gap is an upper bound of the dual suboptimality, $P(w^{(t-1)}) - D(\alpha^{(t-1)}, \beta^{(t-1)}) \ge \epsilon_D^{(t-1)}$. Then (28) implies that

$$\mathbb{E}\left[\frac{\epsilon_D^{(t)}}{n}\right] \le (1-q)\mathbb{E}\left[\frac{\epsilon_D^{(t-1)}}{n}\right] + q^2\frac{G}{2\lambda}, \text{ where } G = \frac{1}{n}\sum_{\ell=1}^m G_\ell = 4RL^2$$

Starting from this recursion, we can now apply the same analysis for $L$-Lipschitz loss functions of the single-machine SDCA in (Shalev-Shwartz and Zhang, 2013) to obtain the following desired inequality:

$$\mathbb{E}\left[\frac{\epsilon_D^{(t)}}{n}\right] \le \frac{2G}{\lambda(2\tilde{n} + t - t_0)}, \tag{29}$$

for all $t \ge t_0 = \max(0, \lceil \tilde{n}\log(\frac{2\lambda\epsilon_D^{(0)}\tilde{n}}{nG})\rceil)$, where $\tilde{n} = \max_\ell(n_\ell/M_\ell)$. Further applying the same strategies in (Shalev-Shwartz and Zhang, 2013) based on (29) proves the desired bound. ■

### 9.8 Proof of Theorem 11

Our proof strategy follows (Shalev-Shwartz and Zhang, 2014) and (Frostig et al., 2015), which both used acceleration techniques of (Nesterov, 2004) on top of approximate proximal point steps, the main differences compared with (Shalev-Shwartz and Zhang, 2014) and (Frostig et al., 2015) are here we warm start with two groups dual variables ($\alpha$ and $\beta$) where (Shalev-Shwartz and Zhang, 2014) warm start only with $\alpha$ as it consider the single machine setting, and (Frostig et al., 2015) warm start from primal variables $w$.

**Proof** The proof consists of the following steps:

- In Lemma 17 we show that one can construct a quadratic lower bound of the original objective $P(w)$ from an approximate minimizer of the proximal objective $P_t(w)$.
- Using the quadratic lower bound we construct an estimation sequence, based on which in Lemma 18 we prove the accelerated convergence rate for the outer loops.
- We show in Lemma 19 that by warm start the iterates from the last stage, the dual sub-optimality for the next stage is small.

Based on Lemma 19, we know the contraction factor between the initial dual sub-optimality and the target primal-dual gap at stage $t$ can be upper bounded by

$$\begin{aligned}\frac{D_t(\alpha_{\text{opt}}^{(t)}, \beta_{\text{opt}}^{(t)}) - D_t(\alpha^{(t-1)}, \beta^{(t-1)})}{(\eta\xi_{t-1})/(2+2\eta^{-2})} \leq& \frac{\epsilon_{t-1}}{(\eta\xi_{t-1})/(2+2\eta^{-2})} + \frac{36\kappa\xi_{t-3}}{\lambda(\eta\xi_{t-1})/(2+2\eta^{-2})} \\ \leq& \frac{1}{1-\eta/2} + \frac{36(2+2\eta^{-2})}{\eta(1-\eta/2)^2} \cdot \frac{\kappa}{\lambda} \\ \leq& 2 + \frac{36}{\eta^5(1-\eta^2)}\end{aligned}$$

where the last step we used the fact that $\eta^{-4} = (\eta^{-2}-1)(\eta^{-2}+1)+1 > (\eta^{-2}-1)(\eta^{-2}+1) = \frac{2\kappa(\eta^{-2}+1)}{\lambda}$. Thus using the results from plain DADM (Theorem 6), we know the number of inner iterations in each stage is upper bounded by

$$\begin{aligned}&\chi\left(\log(\chi) + \log\left(\frac{D_t(\alpha_{\text{opt}}^{(t)}, \beta_{\text{opt}}^{(t)}) - D_t(\alpha^{(t-1)}, \beta^{(t-1)})}{(\eta\xi_{t-1})/(2+2\eta^{-2})}\right)\right) \\ \leq& \chi\left(\log(\chi) + 7 + \frac{5}{2}\log\left(\frac{\lambda+2\kappa}{\lambda}\right)\right),\end{aligned}$$

where $\chi = \frac{R}{\gamma(\lambda+\kappa)} + \max_\ell \frac{n_\ell}{M_\ell}$. ■

### 9.9 Proof of Corollary 13

By the property of $\tilde{\phi}_i(u_i)$, for every $w$ we have

$$0 \leq \frac{\hat{P}(w)}{n} - \frac{P(w)}{n} \leq \frac{\gamma L^2}{2},$$

thus if we found a predictor $w^{(t)}$ that is $\frac{\epsilon}{2}$-suboptimal with respect to $\frac{\hat{P}(w)}{n}$:

$$\frac{\hat{P}(w^{(t)})}{n} - \min_w \frac{\hat{P}(w)}{n} \le \frac{\epsilon}{2},$$

and we choose $\gamma = \epsilon / L^2$, we know it must be $\epsilon$-suboptimal with respect to $\frac{P(w)}{n}$, because

$$\begin{aligned}\frac{P(w^{(t)})}{n} - \frac{P(w^*)}{n} \le& \frac{\hat{P}(w^{(t)})}{n} - \frac{\hat{P}(w^*)}{n} + \frac{\gamma L^2}{2} \\ \le& \frac{\hat{P}(w^{(t)})}{n} - \min_w \frac{\hat{P}(w)}{n} + \frac{\epsilon}{2} \le \epsilon.\end{aligned}$$

The rest of the proof just follows the smooth case as proved in Theorem 11.

**Dual subproblems in Acc-DADM** Define: $\tilde{\lambda} = \lambda + \kappa$, $f(w) = \frac{\lambda}{\tilde{\lambda}} g(w) + \frac{\kappa}{2\tilde{\lambda}} \|w\|_2^2$. Let

$$\begin{aligned}P_t(w) =& \sum_{i=1}^n \phi_i(X_i^\top w) + \lambda n g(w) + h(w) + \frac{\kappa n}{2} \left\|w - y^{(t-1)}\right\|_2^2 \\ =& \sum_{i=1}^n \phi_i(X_i^\top w) + \tilde{\lambda} n \left( f(w) - \frac{\kappa}{\tilde{\lambda}} w^\top y^{(t-1)} \right) + h(w) + \frac{\kappa n_\ell}{2} \left\|y^{(t-1)}\right\|_2^2\end{aligned}$$

be the global primal problem to solve, and

$$P_{\ell_t}(w_\ell) = \sum_{i \in S_\ell} \phi_i(X_i^\top w_\ell) + \lambda n_\ell g(w_\ell) + \frac{\kappa n_\ell}{2} \left\|w_\ell - y^{(t-1)}\right\|_2^2$$

be the separated local problem. Given each dual variable $\beta_\ell$, we also define the adjusted local primal problem as:

$$\tilde{P}_{\ell_t}(w_\ell | \beta_\ell) = \sum_{i \in S_\ell} \phi_i(X_i^\top w_\ell) + \lambda n_\ell g(w_\ell) + \beta_\ell^\top w_\ell + \frac{\kappa n_\ell}{2} \left\|w_\ell - y^{(t-1)}\right\|_2^2,$$

it is not hard to see the adjusted local dual problem is

$$\tilde{D}_{\ell_t}(\alpha_{(\ell)} | \beta_\ell) = \sum_{i \in S_\ell} -\phi_i^*(-\alpha_i) - \tilde{\lambda} n_\ell f^* \left( \frac{\sum_{i \in S_\ell} X_i \alpha_i - \beta_\ell + \kappa n_\ell y^{(t-1)}}{\tilde{\lambda} n_\ell} \right) + \frac{\kappa n_\ell}{2} \left\|y^{(t-1)}\right\|_2^2,$$

and the global dual objective can be written as

$$D_t(\alpha, \beta) = \sum_{\ell=1}^m \tilde{D}_{\ell_t}(\alpha_{(\ell)} | \beta_\ell) - h^* \left( \sum_{\ell=1}^m \beta_\ell \right).$$

**Quadratic lower bound for $P(w)$ based on approximate proximal point algorithm** Since $P_t(w) = P(w) + \frac{\kappa n}{2} \left\|w - y^{(t-1)}\right\|_2^2$, and let $w_{\text{opt}}^{(t)} = \arg\min_w P_t(w)$. The following lemma shows we could construct a lower bound of $P(w)$ from an approximate minimizer of $P_t(w)$.

**Lemma 17** *Let $w^+$ be an $\epsilon$-approximated minimizer of $P_t(w)$, i.e.*

$$P_t(w^+) \leq P_t(w_{\text{opt}}^{(t)}) + \epsilon.$$

*We can construct the following quadratic lower bound for $P(w)$, as $\forall w$*

$$P(w) \geq P(w^+) + Q(w; w^+, y^{(t-1)}, \epsilon), \tag{30}$$

*where*

$$\begin{aligned} Q(w; w^+, y^{(t-1)}, \epsilon) =& \frac{\lambda n}{4} \left\| w - \left( y^{(t-1)} - \left(1 + \frac{2\kappa}{\lambda}\right) (y^{(t-1)} - w^+) \right) \right\|_2^2 \\ & - \frac{\kappa^2 n}{\lambda} \left\| w^+ - y^{(t-1)} \right\|_2^2 - \left( \frac{2\kappa + 2\lambda}{\lambda} \right) \epsilon. \end{aligned}$$

**Proof** Since $w_{\text{opt}}^{(t)}$ is the minimizer of a $(\kappa + \lambda)n$-strongly convex objective $P_t(w)$, we know $\forall w$,

$$\begin{aligned} P_t(w) \geq & P_t(w_{\text{opt}}^{(t)}) + \frac{(\kappa + \lambda)n}{2} \left\| w - w_{\text{opt}}^{(t)} \right\|_2^2 \\ \geq & P_t(w^+) + \frac{(\kappa + \lambda)n}{2} \left\| w - w_{\text{opt}}^{(t)} \right\|_2^2 - \epsilon, \end{aligned}$$

which is equivalent to

$$P(w) \geq P(w^+) + \frac{(\kappa + \lambda)n}{2} \left\| w - w_{\text{opt}}^{(t)} \right\|_2^2 - \epsilon + \frac{\kappa n}{2} \left( \left\| w^+ - y^{(t-1)} \right\|_2^2 - \left\| w - y^{(t-1)} \right\|_2^2 \right).$$

Since

$$\begin{aligned} \frac{\kappa + \lambda/2}{2} \left\| w - w^+ \right\|_2^2 =& \frac{\kappa + \lambda/2}{2} \left\| w - w_{\text{opt}}^{(t)} + w_{\text{opt}}^{(t)} - w^+ \right\|_2^2 \\ =& \frac{\kappa + \lambda/2}{2} \left( \left\| w - w_{\text{opt}}^{(t)} \right\|_2^2 + \left\| w_{\text{opt}}^{(t)} - w^+ \right\|_2^2 \right) \\ & + (\kappa + \lambda/2) \langle w - w_{\text{opt}}^{(t)}, w_{\text{opt}}^{(t)} - w^+ \rangle \\ \leq & \frac{\kappa + \lambda/2}{2} \left( \left\| w - w_{\text{opt}}^{(t)} \right\|_2^2 + \left\| w_{\text{opt}}^{(t)} - w^+ \right\|_2^2 \right) + \frac{\lambda/2}{2} \left\| w_{\text{opt}}^{(t)} - w \right\|_2^2 \\ & + \frac{(\kappa + \lambda/2)^2}{\lambda} \left\| w^+ - w_{\text{opt}}^{(t)} \right\|_2^2, \end{aligned}$$

re-organizing terms we get

$$\frac{\kappa + \lambda}{2} \left\| w_{\text{opt}}^{(t)} - w \right\|_2^2 \geq \frac{\kappa + \lambda/2}{2} \left\| w - w^+ \right\|_2^2 - \frac{(\kappa + \lambda)(\kappa + \lambda/2)}{\lambda} \left\| w^+ - w_{\text{opt}}^{(t)} \right\|_2^2$$

So

$$\begin{aligned} P(w) \geq & P(w^+) + \frac{(\kappa + \lambda/2)n}{2} \left\| w - w^+ \right\|_2^2 - \frac{(\kappa + \lambda)(\kappa + \lambda/2)n}{\lambda} \left\| w^+ - w_{\text{opt}}^{(t)} \right\|_2^2 - \epsilon \\ & + \frac{\kappa n}{2} \left( \left\| w^+ - y^{(t-1)} \right\|_2^2 - \left\| w - y^{(t-1)} \right\|_2^2 \right) \end{aligned}$$

Also noted that $\frac{(\kappa+\lambda)n}{2} \left\| w^+ - w_{\text{opt}}^{(t)} \right\|_2^2 \leq \epsilon$, we get

$$\begin{aligned} P(w) \geq & P(w^+) + \frac{(\kappa + \lambda/2)n}{2} \left\| w - w^+ \right\|_2^2 - \left( \frac{2\kappa + 2\lambda}{\lambda} \right) \epsilon \\ & + \frac{\kappa n}{2} \left( \left\| w^+ - y^{(t-1)} \right\|_2^2 - \left\| w - y^{(t-1)} \right\|_2^2 \right) \end{aligned}$$

Decompose $\left\| w - w^+ \right\|_2^2$ we get

$$\left\| w - w^+ \right\|_2^2 = \left\| w - y^{(t-1)} \right\|^2 + \left\| y^{(t-1)} - w^+ \right\|_2^2 + 2\langle w - y^{(t-1)}, y^{(t-1)} - w^+ \rangle.$$

So

$$\begin{aligned} P(w) \geq & P(w^+) + \frac{(\lambda/2)n}{2} \left\| w - y^{(t-1)} \right\|_2^2 - \left( \frac{2\kappa + 2\lambda}{\lambda} \right) \epsilon \\ & + \frac{(2\kappa + \lambda/2)n}{2} \left\| y^{(t-1)} - w^+ \right\|_2^2 + (\kappa + \lambda/2)n \langle w - y^{(t-1)}, y^{(t-1)} - w^+ \rangle \end{aligned}$$

Noticed that the right hand side of above inequality is a quadratic function with respect to $w$, and the minimum is achieved when

$$w = y^{(t-1)} - \left( 1 + \frac{2\kappa}{\lambda} \right) (y^{(t-1)} - w^+),$$

with minimum value

$$-\frac{\kappa^2 n}{\lambda} \left\| w^+ - y^{(t-1)} \right\|_2^2 - \left( \frac{2\kappa + 2\lambda}{\lambda} \right) \epsilon,$$

with above we finished the proof of Lemma 17. ■

**Convergence proof** Define the following sequence of quadratic functions

$$\psi_0(w) = P(0) + \frac{\lambda n}{4} \|w\|_2^2 - \left( \frac{2\kappa + 2\lambda}{\lambda} \right) (P(0) - D(0, 0)),$$

and for $t \geq 1$,

$$\psi_t(w) = (1 - \eta)\psi_{t-1}(w) + \eta(P(w^{(t)}) + Q(w; w^{(t)}, y^{(t-1)}, \epsilon_t)),$$

where $\eta = \sqrt{\frac{\lambda}{\lambda + 2\kappa}}$, We first calculate the explicit form of the quadratic function $\psi_t(w)$ and its minimizer $v^{(t)} = \arg\min_w \psi_t(w)$. Clearly $v^{(0)} = 0$, and noticed that $\psi_t(w)$ is always a $\frac{\lambda n}{2}$-strongly convex function, we know $\psi_t(w)$ is in the following form:

$$\psi_t(w) = \psi_t(v^{(t)}) + \frac{\lambda n}{4} \left\| w - v^{(t)} \right\|_2^2.$$

Based on the definition of $\psi_{t+1}(w)$, and $v^{(t+1)}$ is minimizing $\psi_{t+1}(w)$, based on first-order optimality condition, we know

$$\frac{(1-\eta)\lambda n}{2}(v^{(t+1)} - v^{(t)}) + \frac{\eta\lambda n}{2}\left(v^{(t+1)} - \left(y^{(t)} - \left(1+\frac{2\kappa}{\lambda}\right)(y^{(t)} - w^{(t+1)})\right)\right) = 0,$$

rearranging we get

$$v^{(t+1)} = (1-\eta)v^{(t)} + \eta\left(y^{(t)} - \left(1+\frac{2\kappa}{\lambda}\right)(y^{(t)} - w^{(t+1)})\right).$$

The following lemma proves the convergence rate of $w^{(t)}$ to its minimizer.

**Lemma 18** *Let*

$$\epsilon_t \leq \frac{\eta}{2(1+\eta^{-2})}\xi_t,$$

*and*

$$\xi_t = (1-\eta/2)^t \xi_0,$$

*we will have the following convergence guarantee:*

$$P(w^{(t)}) - P(w^*) \leq \xi_t.$$

**Proof** It is sufficient to prove

$$P(w^{(t)}) - \min_w \psi_t(w) \leq \xi_t, \tag{31}$$

then we get

$$P(w^{(t)}) - P(w^*) \leq P(w^{(t)}) - \psi_t(w^*) \leq P(w^{(t)}) - \min_w \psi_t(w) \leq \xi_t.$$

We prove equation (31) by induction. When $t = 0$, we have

$$P(w^{(0)}) - \phi_0(v^{(0)}) = \left(\frac{2\kappa + 2\lambda}{\lambda}\right)\epsilon_0 = \xi_0,$$

which verified (31) is true for $t = 0$. Suppose the claim holds for some $t \geq 1$, for the stage $t+1$, we have

$$\begin{aligned}
\psi_{t+1}(v^{(t+1)}) =&(1-\eta)\left(\psi_t(v^{(t)}) + \frac{\lambda n}{4}\left\|v^{(t+1)} - v^{(t)}\right\|_2^2\right) \\
&+ \eta(P(w^{(t+1)}) + Q(v^{(t+1)}; w^{(t+1)}, y^{(t)}, \epsilon_t)) \\
=&(1-\eta)\psi_t(v^{(t)}) + \frac{(1-\eta)\eta^2\lambda n}{4}\left\|v^{(t)} - \left(y^{(t)} - \left(1+\frac{2\kappa}{\lambda}\right)(y^{(t)} - w^{t+1})\right)\right\|_2^2 \\
&+ \eta P(w^{(t+1)}) + \frac{\eta(1-\eta)^2\lambda n}{4}\left\|v^{(t)} - \left(y^{(t)} - \left(1+\frac{2\kappa}{\lambda}\right)(y^{(t)} - w^{(t+1)})\right)\right\|_2^2 \\
&- \frac{\eta\kappa^2 n}{\lambda}\left\|y^{(t)} - w^{(t+1)}\right\|_2^2 - \eta\left(\frac{2\kappa+2\lambda}{\lambda}\right)\epsilon_t \\
=&(1-\eta)\psi_t(v^{(t)}) + \eta P(w^{(t+1)}) - \frac{\eta\kappa^2 n}{\lambda}\left\|y^{(t)} - w^{(t+1)}\right\|_2^2 - \eta\left(\frac{2\kappa+2\lambda}{\lambda}\right)\epsilon_t \\
&+ \frac{\eta(1-\eta)\lambda n}{4}\left\|v^{(t)} - \left(y^{(t)} - \left(1+\frac{2\kappa}{\lambda}\right)(y^{(t)} - w^{(t+1)})\right)\right\|_2^2.
\end{aligned}$$

Since

$$\begin{aligned}
&- \frac{\eta\kappa^2 n}{\lambda}\left\|y^{(t)} - w^{(t+1)}\right\|_2^2 + \frac{\eta(1-\eta)\lambda n}{4}\left\|v^{(t)} - \left(y^{(t)} - \left(1+\frac{2\kappa}{\lambda}\right)(y^{(t)} - w^{(t+1)})\right)\right\|_2^2 \\
\geq&\left(-\frac{\eta\kappa^2 n}{\lambda} + \frac{\eta(1-\eta)\lambda n}{4}\left(1+\frac{2\kappa}{\lambda}\right)^2\right)\left\|y^{(t)} - w^{(t+1)}\right\|_2^2 \\
&+ \eta(1-\eta)n\left(\kappa+\frac{\lambda}{2}\right)\langle v^{(t)} - y^{(t)}, y^{(t)} - w^{(t+1)}\rangle \\
\geq&\left(-\frac{\eta\kappa^2 n}{\lambda} + \frac{\eta(1-\eta)\kappa^2 n}{\lambda}\right)\left\|y^{(t)} - w^{(t+1)}\right\|_2^2 \\
&+ \eta(1-\eta)n\left(\kappa+\frac{\lambda}{2}\right)\langle v^{(t)} - y^{(t)}, y^{(t)} - w^{(t+1)}\rangle \\
=&-\frac{\eta^2\kappa^2 n}{\lambda}\left\|y^{(t)} - w^{(t+1)}\right\|_2^2 + \eta(1-\eta)n\left(\kappa+\frac{\lambda}{2}\right)\langle v^{(t)} - y^{(t)}, y^{(t)} - w^{(t+1)}\rangle
\end{aligned}$$

Thus

$$\begin{aligned}
\psi_{t+1}(v^{(t+1)}) \geq& (1-\eta)\psi_t(v^{(t)}) + \eta P(w^{(t+1)}) - \eta\left(\frac{2\kappa+2\lambda}{\lambda}\right)\epsilon_t \\
&- \frac{\eta^2\kappa^2 n}{\lambda}\left\|y^{(t)} - w^{(t+1)}\right\|_2^2 + \eta(1-\eta)n\left(\kappa+\frac{\lambda}{2}\right)\langle v^{(t)} - y^{(t)}, y^{(t)} - w^{(t+1)}\rangle
\end{aligned}$$

Also using (30) with $w = w^{(t)}$, we have

$$\begin{aligned}
P(w^{(t)}) \geq & P(w^{(t+1)}) + \frac{\lambda n}{4} \left\| w^{(t)} - \left( y^{(t)} - \left(1 + \frac{2\kappa}{\lambda}\right) (y^{(t)} - w^{(t+1)}) \right) \right\|_2^2 \\
& - \frac{\kappa^2 n}{\lambda} \left\| w^{(t+1)} - y^{(t)} \right\|_2^2 - \left( \frac{2\kappa + 2\lambda}{\lambda} \right) \epsilon_t \\
\geq & P(w^{(t+1)}) + n \left( \kappa + \frac{\lambda}{2} \right) \langle w^{(t)} - y^{(t)}, y^{(t)} - w^{(t+1)} \rangle - \left( \frac{2\kappa + 2\lambda}{\lambda} \right) \epsilon_t \\
& + \left( \frac{\lambda n}{4} \left( 1 + \frac{2\kappa}{\lambda} \right)^2 - \frac{\kappa^2 n}{\lambda} \right) \left\| w^{(t+1)} - y^{(t)} \right\|_2^2 \\
\geq & P(w^{(t+1)}) + n \left( \kappa + \frac{\lambda}{2} \right) \langle w^{(t)} - y^{(t)}, y^{(t)} - w^{(t+1)} \rangle - \left( \frac{2\kappa + 2\lambda}{\lambda} \right) \epsilon_t \\
& + \kappa n \left\| w^{(t+1)} - y^{(t)} \right\|_2^2 .
\end{aligned}$$

we get

$$\begin{aligned}
P(w^{(t+1)}) - \psi_{t+1}(v^{(t+1)}) \leq & (1-\eta)(P(w^{(t)}) - \psi_t(v^{(t)})) + \left( \frac{2\kappa + 2\lambda}{\lambda} \right) \epsilon_t \\
& + \left( \frac{\eta^2 \kappa^2 n}{\lambda} - (1-\eta)\kappa n \right) \left\| y^{(t)} - w^{(t+1)} \right\|_2^2 \\
& + (1-\eta) n \left( \kappa + \frac{\lambda}{2} \right) \langle y^{(t)} - w^{(t)}, y^{(t)} - w^{(t+1)} \rangle . \\
& + \eta(1-\eta) n \left( \kappa + \frac{\lambda}{2} \right) \langle y^{(t)} - v^{(t)}, y^{(t)} - w^{(t+1)} \rangle
\end{aligned}$$

Since

$$\begin{aligned}
(1-\eta)(P(w^{(t)}) - \psi_t(v^{(t)})) + \left( \frac{2\kappa + 2\lambda}{\lambda} \right) \epsilon_t \leq & (1-\eta)\xi_t + \left( \frac{2\kappa + 2\lambda}{\lambda} \right) \cdot \frac{\eta}{2(1+\eta^{-2})} \xi_t \\
= & \left( 1 - \eta + \frac{\eta}{2} \right) \xi_t = \xi_{t+1},
\end{aligned}$$

and

$$\frac{\eta^2 \kappa^2 n}{\lambda} - (1-\eta)\kappa n = \kappa n \left( \frac{\kappa}{\lambda + 2\kappa} + \sqrt{\frac{\lambda}{\lambda + 2\kappa}} - 1 \right) \leq 0,$$

If we set $y^{(t)} = (\eta v^{(t)} + w^{(t)})/(1+\eta)$, which is equivalent to the update rule as $y^{(t)} = w^{(t)} + \nu(w^{(t)} - w^{(t-1)})$, because in that way we have

$$\begin{aligned}
\eta v^{(t)} = & \eta((1-\eta)v^{(t-1)}) + \eta^2 \left( y^{(t-1)} - \left( 1 + \frac{2\kappa}{\lambda} \right) (y^{(t-1)} - w^{(t)}) \right) \\
= & w^{(t)} + (1-\eta)(\eta v^{(t-1)} - (1+\eta) y^{(t-1)}) \\
= & w^{(t)} - (1-\eta) w^{(t-1)},
\end{aligned}$$

thus

$$\begin{aligned}y^{(t)} =&\frac{\eta v^{(t)} + w^{(t)}}{1+\eta}\\ =&\frac{2w^{(t)} - (1-\eta)w^{(t-1)}}{1+\eta}\\ =&w^{(t)} + \nu(w^{(t)} - w^{(t-1)}).\end{aligned}$$

So

$$y^{(t)} - w^{(t)} + \eta(y^{(t)} - v^{(t)}) = 0,$$

combining above we obtain

$$P(w^{(t+1)}) - \psi_{t+1}(v^{(t+1)}) \le \xi_{t+1},$$

which concludes the proof. ∎

**Initial dual sub-optimality in each acceleration stage** In the lemma below we upper bound the quantity

$$D_t(\alpha_{\text{opt}}^{(t)}, \beta_{\text{opt}}^{(t)}) - D_t(\alpha^{(t-1)}, \beta^{(t-1)}),$$

where $\alpha_{\text{opt}}^{(t)}, \beta_{\text{opt}}^{(t)} = \arg\max_{\alpha,\beta} D_t(\alpha, \beta)$.

**Lemma 19** *We have the following upper bound on the initial dual sub-optimality at stage $t$:*

$$D_t(\alpha_{\text{opt}}^{(t)}, \beta_{\text{opt}}^{(t)}) - D_t(\alpha^{(t-1)}, \beta^{(t-1)}) \le \epsilon_{t-1} + \frac{36\kappa}{\lambda}\xi_{t-3}.$$

**Proof** On one hand, since $f(\cdot)$ is 1-strongly convex, we know $f^*(\cdot)$ is 1-smooth. Thus

$$\begin{aligned}&\tilde{\lambda} n_\ell f^*\left(\frac{\sum_{i\in S_\ell} X_i \alpha_i^{(t-1)} - \beta_\ell^{(t-1)} + \kappa n_\ell y^{(t-1)}}{\tilde{\lambda} n_\ell}\right)\\ \le&\tilde{\lambda} n_\ell f^*\left(\frac{\sum_{i\in S_\ell} X_i \alpha_i^{(t-1)} - \beta_\ell^{(t-1)} + \kappa n_\ell y^{(t-2)}}{\tilde{\lambda} n_\ell}\right)\\ &+ \kappa n_\ell \nabla f^*\left(\frac{\sum_{i\in S_\ell} X_i \alpha_i^{(t-1)} - \beta_\ell^{(t-1)} + \kappa n_\ell y^{(t-2)}}{\tilde{\lambda} n_\ell}\right)^\top (y^{(t-1)} - y^{(t-2)})\\ &+ \frac{\kappa^2 n_\ell^2}{2\tilde{\lambda} n_\ell}\left\|y^{(t-1)} - y^{(t-2)}\right\|_2^2,\end{aligned}$$

noted that

$$\nabla f^*\left(\frac{\sum_{i\in S_\ell} X_i \alpha_i^{(t-1)} - \beta_\ell^{(t-1)} + \kappa n_\ell y^{(t-2)}}{\tilde{\lambda} n_\ell}\right) = w^{(t-1)},$$

we see

$$
\begin{aligned}
& -\tilde{D}_{\ell_t}(\alpha_{(\ell)}^{(t-1)}|\beta_\ell^{(t-1)}) + \tilde{D}_{\ell_{t-1}}(\alpha_{(\ell)}^{(t-1)}|\beta_\ell^{(t-1)}) \\
\leq & \kappa n_\ell {w_\ell^{(t-1)}}^\top (y^{(t-1)} - y^{(t-2)}) + \frac{\kappa^2 n_\ell^2}{2\tilde{\lambda} n_\ell} \left\| y^{(t-1)} - y^{(t-2)} \right\|_2^2 \\
& + \frac{\kappa n_\ell}{2} \left\| y^{(t-2)} \right\|_2^2 - \frac{\kappa n_\ell}{2} \left\| y^{(t-1)} \right\|_2^2 .
\end{aligned}
$$

On the other hand, since

$$
\begin{aligned}
& \tilde{P}_{\ell_t}(w_\ell^{(t-1)}|\beta_\ell^{(t-1)}) - \tilde{P}_{\ell_{t-1}}(w_\ell^{(t-1)}|\beta_\ell^{(t-1)}) \\
= & \kappa n_\ell {w_\ell^{(t-1)}}^\top (y^{(t-2)} - y^{(t-1)}) - \frac{\kappa n_\ell}{2} \left\| y^{(t-2)} \right\|_2^2 + \frac{\kappa n_\ell}{2} \left\| y^{(t-1)} \right\|_2^2 .
\end{aligned}
$$

Combining above we know

$$
\begin{aligned}
& \tilde{P}_{\ell_t}(w_\ell^{(t-1)}|\beta_\ell^{(t-1)}) - \tilde{D}_{\ell_t}(\alpha_{(\ell)}^{(t-1)}|\beta_\ell^{(t-1)}) \\
\leq & \tilde{P}_{\ell_{t-1}}(w_\ell^{(t-1)}|\beta_\ell^{(t-1)}) - \tilde{D}_{\ell_{t-1}}(\alpha_{(\ell)}^{(t-1)}|\beta_\ell^{(t-1)}) + \frac{\kappa^2 n_\ell^2}{2\tilde{\lambda} n_\ell} \left\| y^{(t-1)} - y^{(t-2)} \right\|_2^2 ,
\end{aligned}
$$

Since $\kappa \leq \tilde{\lambda}$, summing over above inequality we know

$$
P_t(w_\ell^{(t-1)}) - D_t(\alpha^{(t-1)}, \beta^{(t-1)}) \leq \epsilon_{t-1} + \frac{\kappa n}{2} \left\| y^{(t-1)} - y^{(t-2)} \right\|_2^2 ,
$$

also noted that $P_t(w_\ell^{(t-1)}) \geq D_t(\alpha_{\text{opt}}^{(t)}, \beta_{\text{opt}}^{(t)})$, we get

$$
D_t(\alpha_{\text{opt}}^{(t)}, \beta_{\text{opt}}^{(t)}) - D_t(\alpha^{(t-1)}, \beta^{(t-1)}) \leq \epsilon_{t-1} + \frac{\kappa n}{2} \left\| y^{(t-1)} - y^{(t-2)} \right\|_2^2 .
$$

For the term $\left\| y^{(t-1)} - y^{(t-2)} \right\|_2^2$, based on the definition of $y^{(t-1)}$ and the fact $\eta \leq 1$, we know

$$
\begin{aligned}
\left\| y^{(t-1)} - y^{(t-2)} \right\|_2 \leq & \left\| w^{(t-1)} - w^{(t-2)} - \eta(w^{(t-1)} - w^{(t-2)} - (w^{(t-2)} - w^{(t-3)})) \right\|_2 \\
\leq & 3 \max_{i=\{1,2\}} \left\| w^{(t-i)} - w^{(t-i-1)} \right\|_2 .
\end{aligned}
$$

Then we upper bound $\left\| w^{(t-i)} - w^{(t-i-1)} \right\|_2$ using objective sub-optimality, using triangle inequality and the fact that $P(w^*)$ is $\lambda n$-strongly convex, we have

$$
\begin{aligned}
\left\| w^{(t-i)} - w^{(t-i-1)} \right\|_2 \leq & \left\| w^{(t-i)} - w^* \right\|_2 + \left\| w^{(t-i-1)} - w^* \right\|_2 \\
\leq & \sqrt{\frac{2(P(w^{(t-i)}) - P(w^*))}{\lambda n}} + \sqrt{\frac{2(P(w^{(t-i-1)}) - P(w^*))}{\lambda n}} \\
\leq & 2\sqrt{2\frac{\xi_{t-i-1}}{\lambda n}} .
\end{aligned}
$$

We know

$$\left\|y^{(t-1)} - y^{(t-2)}\right\|_2^2 \leq 9 \max_{i=\{1,2\}} \left\|w^{(t-i)} - w^{(t-i-1)}\right\|_2^2 \leq \frac{72\xi_{t-3}}{\lambda n}.$$

Combining above we get

$$D_t(\alpha^{(t)}_{\text{opt}}, \beta^{(t)}_{\text{opt}}) - D_t(\alpha^{(t-1)}, \beta^{(t-1)}) \leq \epsilon_{t-1} + \frac{36\kappa}{\lambda}\xi_{t-3}.$$

■

## 10. Experiments

In this section, we apply algorithms to solve $L_2$-$L_1$ regularzied loss minimization problems. We compare **Acc-DADM** to **CoCoA**$^+$ and **OWL-QN** (Andrew and Gao, 2007), as they have already been shown to be superior to other related algorithms in (Yang, 2013; Jaggi et al., 2014; Ma et al., 2017; Andrew and Gao, 2007). For **Acc-DADM** and **CoCoA**$^+$, we apply ProxSDCA of (Shalev-Shwartz and Zhang, 2014) as the local procedure and perform aggressively sequential updates, as the practical variant of DisDCA did in (Yang, 2013) and CoCoA$^+$ did in (Ma et al., 2015). For details about the updates of the local procedure (ProxSDCA), please refer to the application section of (Shalev-Shwartz and Zhang, 2014). For fair comparisons, we use same balanced data partitions and random seeds.

We implement all algorithms using OpenMPI (Graham et al., 2006) and run them on a small cluster inside a private OpenStack cloud service. To simplify the programming efforts, we use one processor to simulate one machine. We test algorithms on four real datasets with different properties (see Table 1). These datasets are publicly available from LIBSVM dataset collections[1].

Table 1: Datasets

| **Dataset** | **Size** ($n$) | **Features** ($d$) | **Sparsity** |
|---|---|---|---|
| *covtype* | 581,012 | 54 | 22.12% |
| *rcv1* | 677,399 | 47,236 | 0.16% |
| *HIGGS* | 11,000,000 | 28 | 92.11% |
| *kdd2010* | 19,264,097 | 29,890,095 | $9.8e^{-7}$ |

**Different loss functions** The optimization problem we consider to solve is

$$\min_w \frac{1}{n}\sum_{i=1}^{n} \phi_i(x_i^\top w) + \frac{\lambda}{2}\|w\|_2^2 + \mu\|w\|_1,$$

where $x_i \in \mathbb{R}^d$ is a feature vector, $y_i \in \{-1, 1\}$ is a binary class label and $\phi_i : \mathbb{R} \to \mathbb{R}$ is the associated loss function. We consider two models: Support Vector Machine (SVM)

1. `https://www.csie.ntu.edu.tw/~cjlin/libsvmtools/datasets`

and Logistic Regression (LR). For SVM, we follow (Shalev-Shwartz and Zhang, 2014) and employ the smooth hinge loss $\tilde{\phi}_i$ (1-smooth) that is:

$$\tilde{\phi}_i(a) = \begin{cases} 0 & a \geq 1 \\ 1 - y_i a - 1/2 & a \leq 0 \\ \frac{1}{2}(1 - y_i a)^2 & o.w. \end{cases} \tag{32}$$

For LR, we employ the logistic loss ($\frac{1}{4}$-smooth) $\phi_i(a) = \log(1 + \exp(-y_i a))$. To apply Acc-DADM, we choose $\lambda g(w) = \frac{\lambda}{2}\|w\|_2^2 + \mu\|w\|_1$ and $h(w) = 0$. Please refer to (Shalev-Shwartz and Zhang, 2014) for detailed derivations of specific loss functions. For datasets with a medium sample size, such as *covtype* and *rcv1*, we employ 8 machines ($m = 8$). For relatively large datasets, such as *HIGGS* and *kdd2010*, we employ 20 machines ($m = 20$). In experiments, we set $\mu = 1e^{-5}$ and vary $\lambda$ in the range $\{1e^{-6}, 1e^{-7}, 1e^{-8}\}$ on different datasets to see the convergence behaviours.

**Mini-batch size** Adjusting the mini-batch size $M_\ell$ corresponds to trading between computation and communication. We use $sp := \frac{M_\ell}{n_\ell}$ to denote the sampling percentage of the local procedure. In experiments, we test three $sp$ values, 0.05 (red), 0.20 (green) and 0.80 (blue). For each case, we run algorithms for 100 passes over the data. As $sp$ increases, the total number of communications needed to run 100 passes through the data decreases.

**Acceleration parameters** For all experiments, we set $\kappa = \frac{mR}{\lambda\gamma} - \lambda$ as our theory suggests. As for $\nu$, our theory suggests $\nu = \frac{1-\eta}{1+\eta}$ where $\eta = \sqrt{\frac{\lambda}{\lambda+2\kappa}}$. While in practice we find that $\nu = 0$ also works well and the algorithm can converge more smoothly. In Figure 1, we plot the empirical convergence results for the theory suggested $\nu$ and the choice $\nu = 0$. We observe that Acc-DADM with the theory suggested $\nu$ does enjoy the acceleration effect, though often converges with rippling behavior. Such a rippling behavior is normal for the accelerated methods, for example, as studied in (Odonoghue and Candes, 2015). Moreover we observe simply setting $\nu = 0$ also works well in practice and produces more smooth convergence behavior, thus we use $\nu = 0$ in later experiments for better visualization.

Figure 2, Figure 3, Figure 4 and Figure 5 show detailed comparison experiments of CoCoA$^+$ and Acc-DADM. Figure 2 and Figure 3 show **the normalized duality gap versus the number of communications** and **the normalized duality gap versus time (s)** experiments for SVM respectively. Figure 4 and Figure 5 show similar plots for LR. From Figure 2 and Figure 4, we can see that larger $M_\ell$ corresponds to more local computations and enjoys less communications accordingly. Looking at Figure 3 and Figure 5 about the normalized duality gap versus time (s), we can see the practical effect of adjusting $M_\ell$. Moreover Figure 6 and Figure 7 show the total comparisons of OWL-QN, CoCoA$^+$ and Acc-DADM by solving LR problems. For OWL-QN, we follow the standard implementation of (Andrew and Gao, 2007) and set the memory parameter as 10. For CoCoA$^+$ and Acc-DADM, we set $sp = 1.0$ and it implies that each communication round corresponds to one pass over the data.

Our empirical studies show that Acc-DADM always yields the best results. When $\lambda$ is relatively large, CoCoA$^+$ sometimes also converges as fast as Acc-DADM. However, CoCoA$^+$ slows down rapidly as $\lambda$ becomes small, while Acc-DADM still enjoys fast convergence. These observations are consistent with our theory.

**Scalability** Scalability is an important metric for distributed algorithms. We study the scalability by observing the number of communications or the running time (s) needed to reach a certain accuracy ($1e^{-3}$ duality gap) versus the number of machines when the mini-batch size is fixed. For balanced partitions, the mini-batch size of each machine is $\frac{n}{m}sp$. In order to fix the mini-batch size, we need to adjust $sp$ accordingly when $m$ varies. In experiments, we vary $sp$ in the range $\{0.04, 0.08, 0.16, 0.32\}$ when $m$ grows exponentially from 4 to 32 or from 5 to 40. For each case, we run algorithms for at most 100 passes over the data. It implies that if the algorithm does not reach enough accuracy within 100 passes over the data, we record the number of communications or the running time as the final value after 100 passes.

Figure 8 and Figure 9 show scalability results when solving SVM problems. Figure 10 and Figure 11 show similar results of LR. From Figure 8 and Figure 10 where y-axis represents the number of communications, we can see the effect of increasing machines only from the algorithm aspect regardless of different communication overheads when employing various distributed computing frameworks (Dean and Ghemawat, 2008; Zaharia et al., 2012; Li et al., 2014). Besides we show the actual running time versus the number of machines in Figure 9 and Figure 11 with the communication time colored green. Empirical results show that Acc-DADM usually enjoys good scalability. Especially when $\lambda$ is relatively small, such as $1e^{-7}$, CoCoA$^+$ may not reach enough accuracy even after 100 passes over the data, while Acc-DADM works significantly better. These observations are consistent with our theory.

**Non-smooth Losses** Our acceleration technique also works for non-smooth losses (see Section 8.2). Figure 12 and Figure 13 show experimental results when employing the hinge loss $\phi_i(a) = \max\{0, 1-a\}$. We can observe that Acc-DADM also enjoys the acceleration effect as in the smooth loss case and converges significantly faster than CoCoA$^+$ especially when $\lambda$ is small.

## 11. Conclusions

In this paper, we have introduced a novel distributed dual formulation for regularized loss minimization problems. Based on this new formulation, we studied a distributed generalization of the single-machine ProxSDCA, which we refer to as DADM. We have shown that the analysis of ProxSDCA can be easily generalized to establish the convergence of DADM. Moreover, we have adapted AccProxSDCA to the distributed setting by using this new dual formulation and provided corresponding theoretical guarantees. We performed numerous experiments on real datasets to validate our theory and show that our new approach improves previous state-of-the-arts in distributed dual optimization.

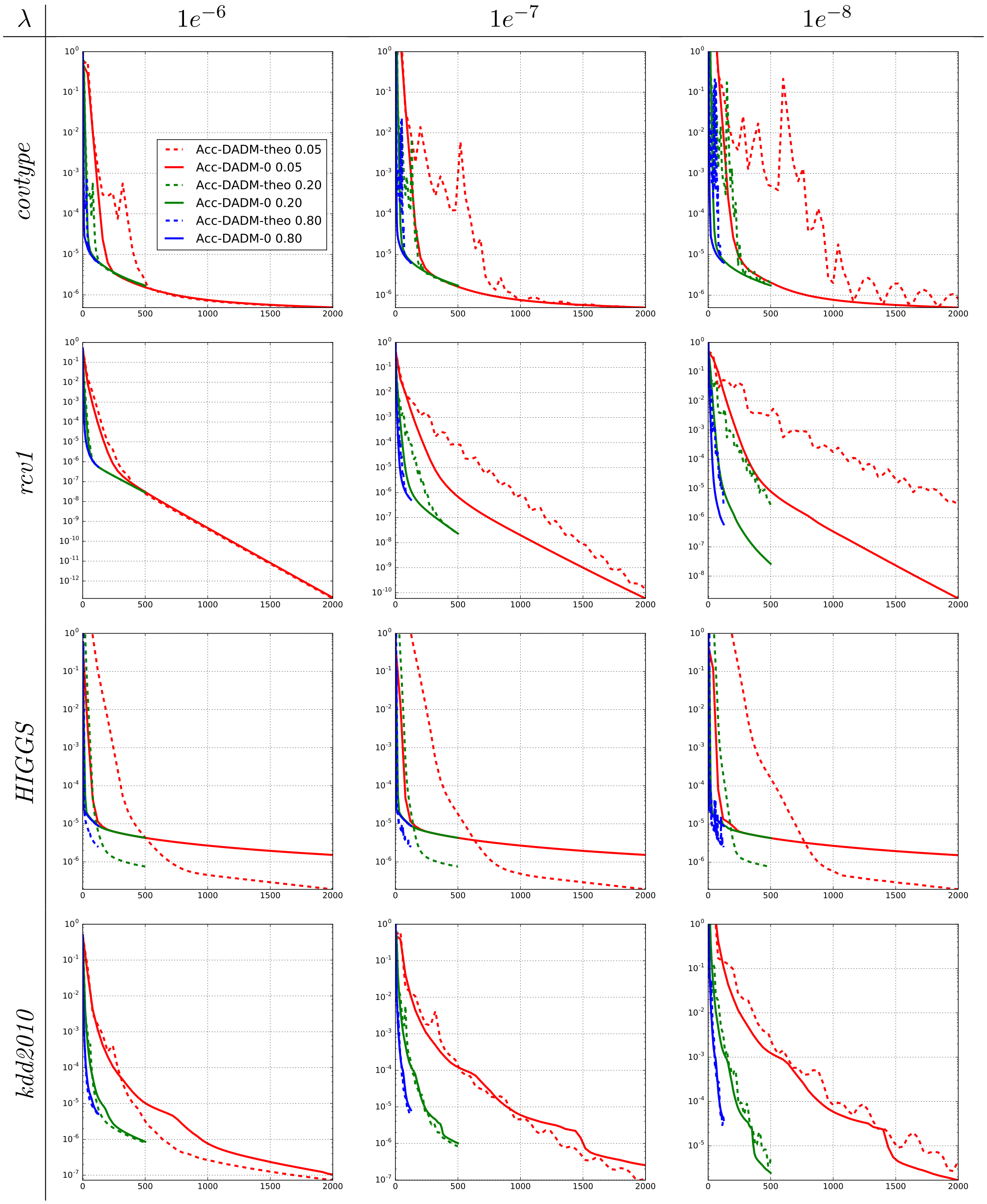

Figure 1: **The normalized duality gap versus the number of communications** of **SVM** experiments with $\mu = 1e^{-5}$, $\lambda$ varing in the range $\{1e^{-6}, 1e^{-7}, 1e^{-8}\}$ and *sp* varing in the range $\{0.05, 0.20, 0.80\}$ on four datasets. Acc-DADM-theo represents the theory suggested $\nu$, Acc-DADM-0 represents the empirical choice $\nu = 0$. We run methods in each case for 100 passes over the data.

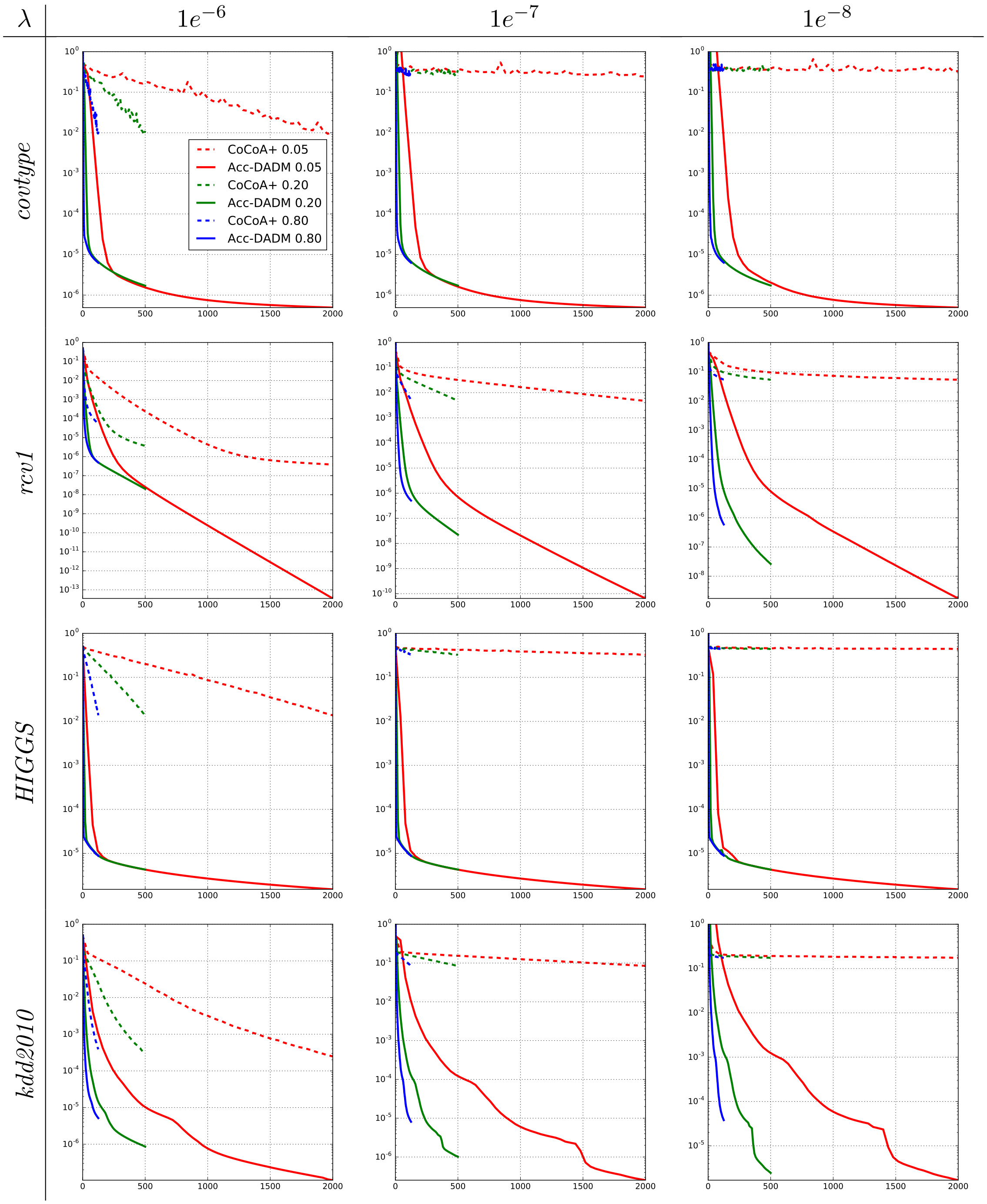

Figure 2: **The normalized duality gap versus the number of communications** of **SVM** experiments with $\mu = 1e^{-5}$, $\lambda$ varing in the range $\{1e^{-6}, 1e^{-7}, 1e^{-8}\}$ and $sp$ varing in the range $\{0.05, 0.20, 0.80\}$ on four datasets. We run methods in each case for 100 passes over the data.

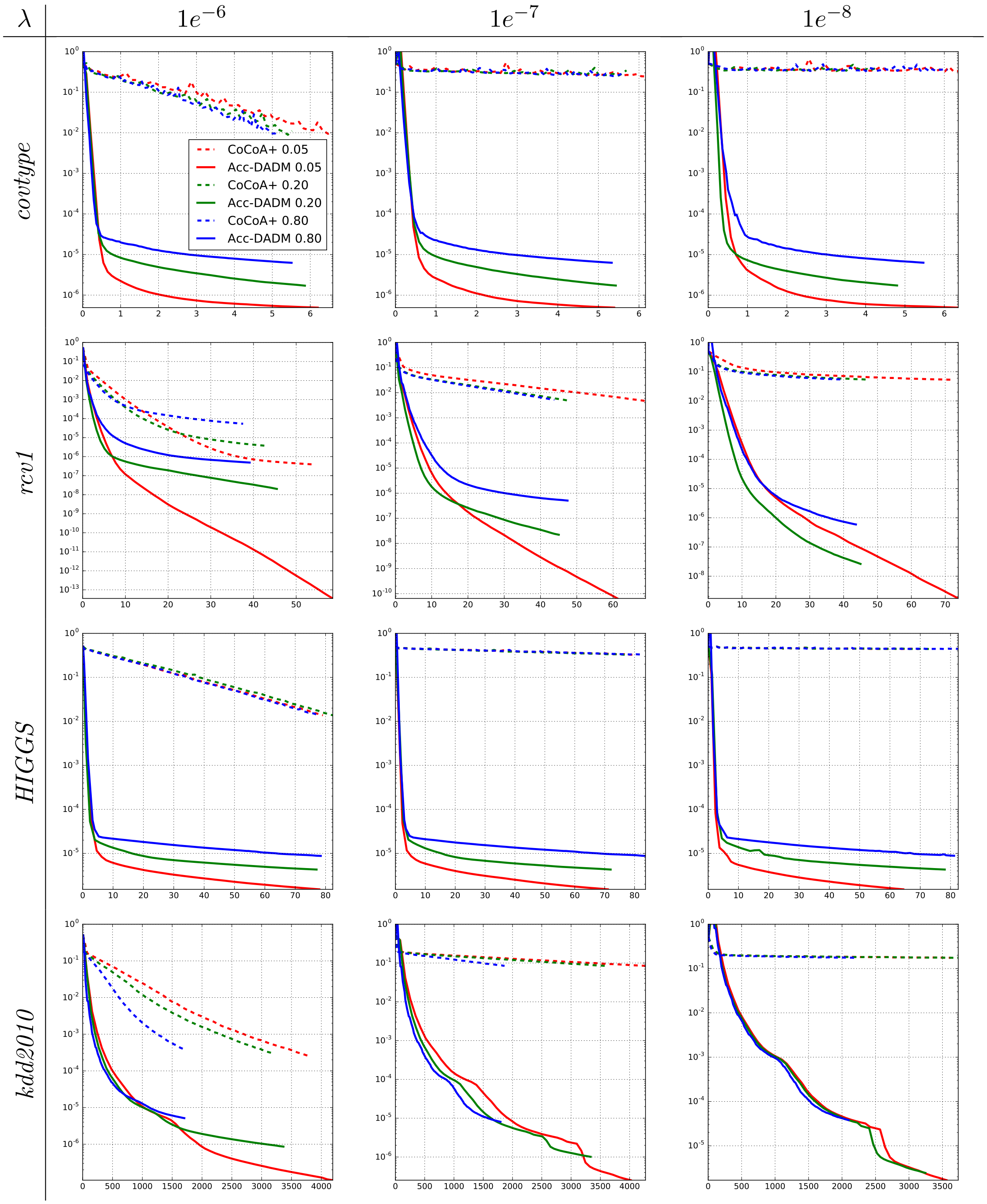

Figure 3: **The normalized duality gap versus time (s)** of **SVM** experiments with $\mu = 1e^{-5}$, $\lambda$ varing in the range $\{1e^{-6}, 1e^{-7}, 1e^{-8}\}$ and $sp$ varing in the range $\{0.05, 0.20, 0.80\}$ on four datasets. We run methods in each case for 100 passes over the data.

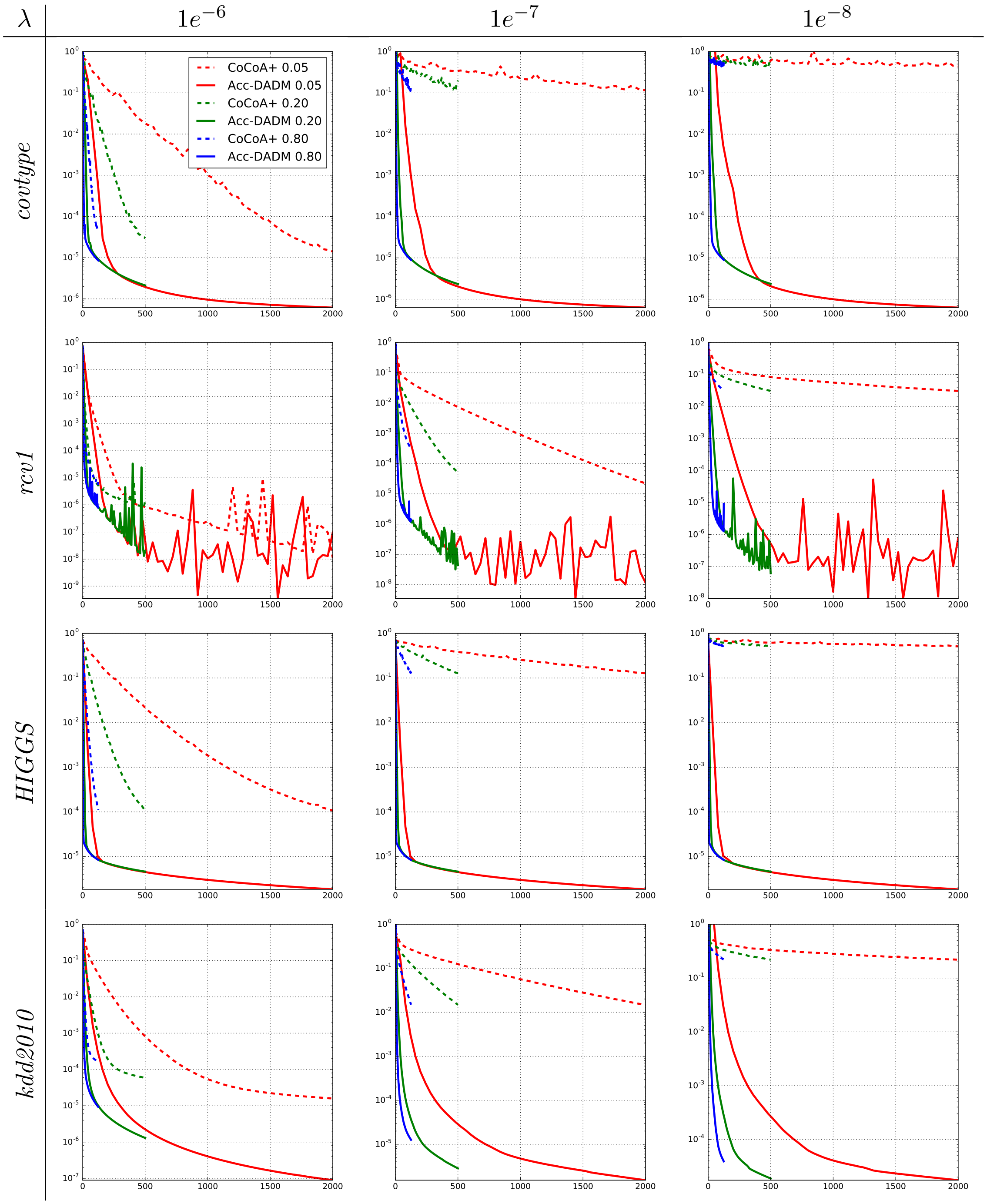

Figure 4: **The normalized duality gap versus the number of communications** of **LR** experiments with $\mu = 1e^{-5}$, $\lambda$ varing in the range $\{1e^{-6}, 1e^{-7}, 1e^{-8}\}$ and *sp* varing in the range $\{0.05, 0.20, 0.80\}$ on four datasets. We run methods in each case for 100 passes over the data.

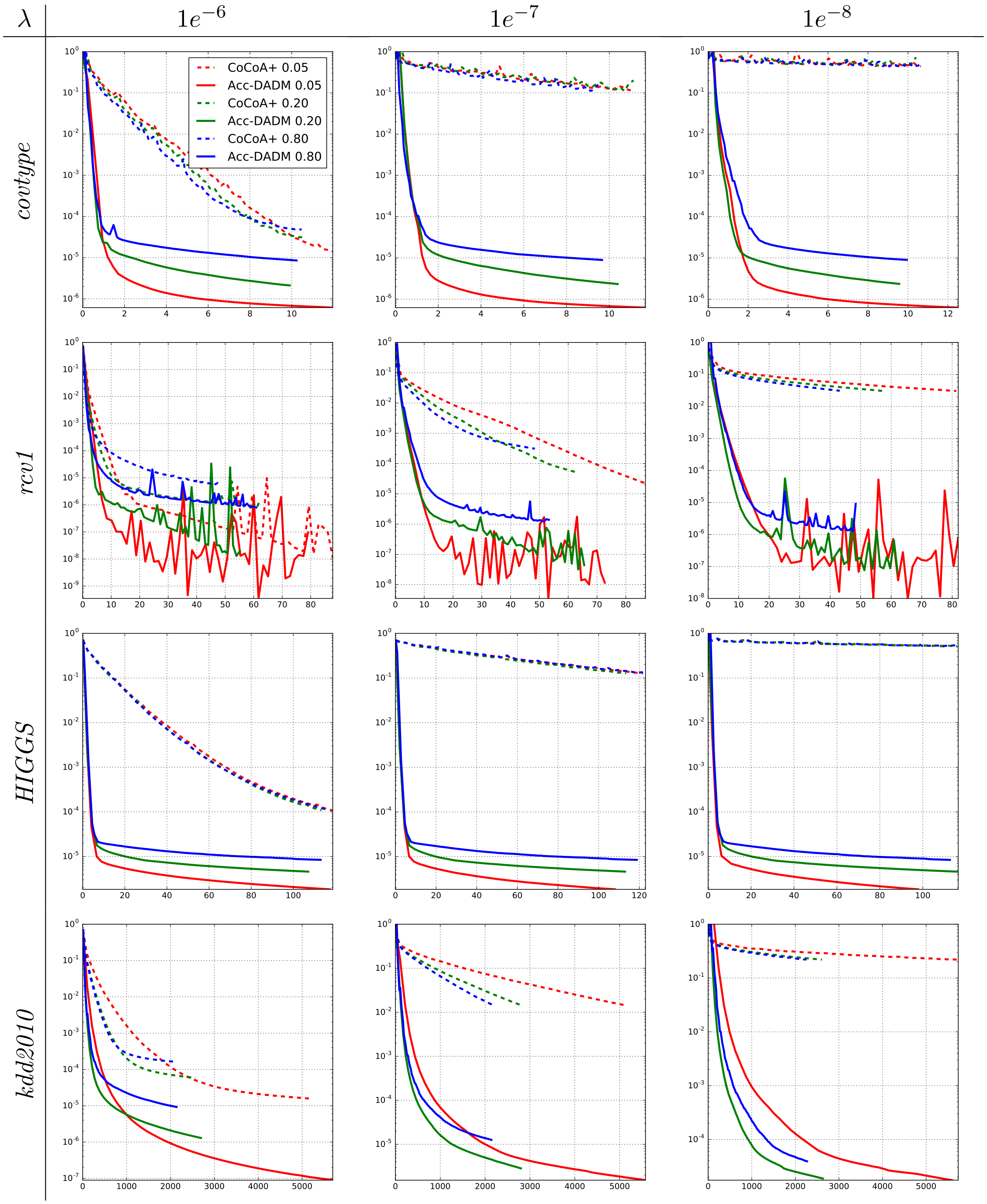

Figure 5: **The normalized duality gap versus time(s)** of LR experiments with $\mu = 1e^{-5}$, $\lambda$ varing in the range $\{1e^{-6}, 1e^{-7}, 1e^{-8}\}$ and $sp$ varing in the range $\{0.05, 0.20, 0.80\}$ on four datasets. We run methods in each case for 100 passes over the data.

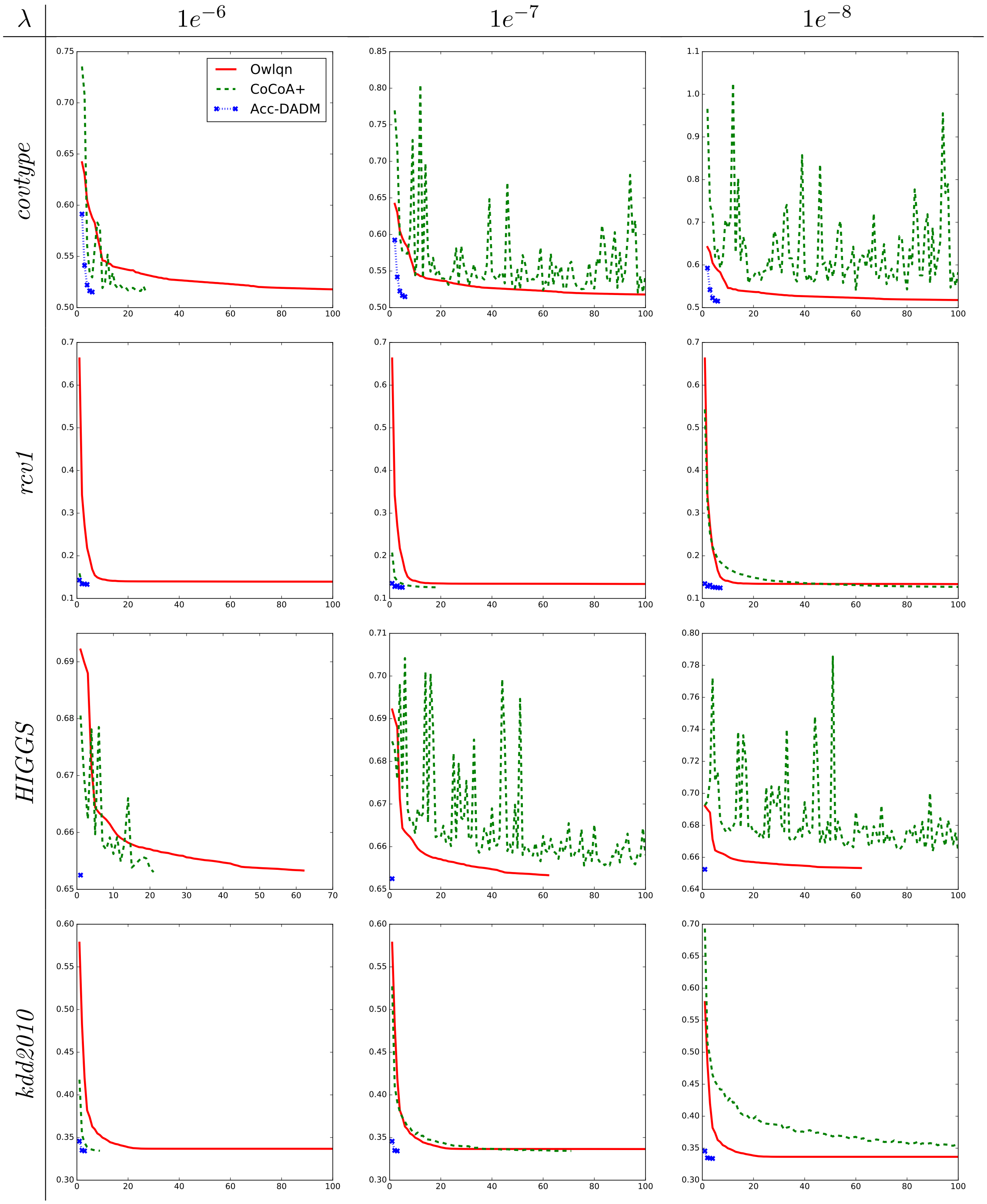

Figure 6: **The normalized primal objective versus the number of passes over the data** of **LR** experiments with $\mu = 1e^{-5}$, $\lambda$ varing in the range $\{1e^{-6}, 1e^{-7}, 1e^{-8}\}$ and $sp = 1.0$ on four datasets. We terminate methods either if the stopping condition is met ($1e^{-3}$ duality gap) or after 100 passes over the data.

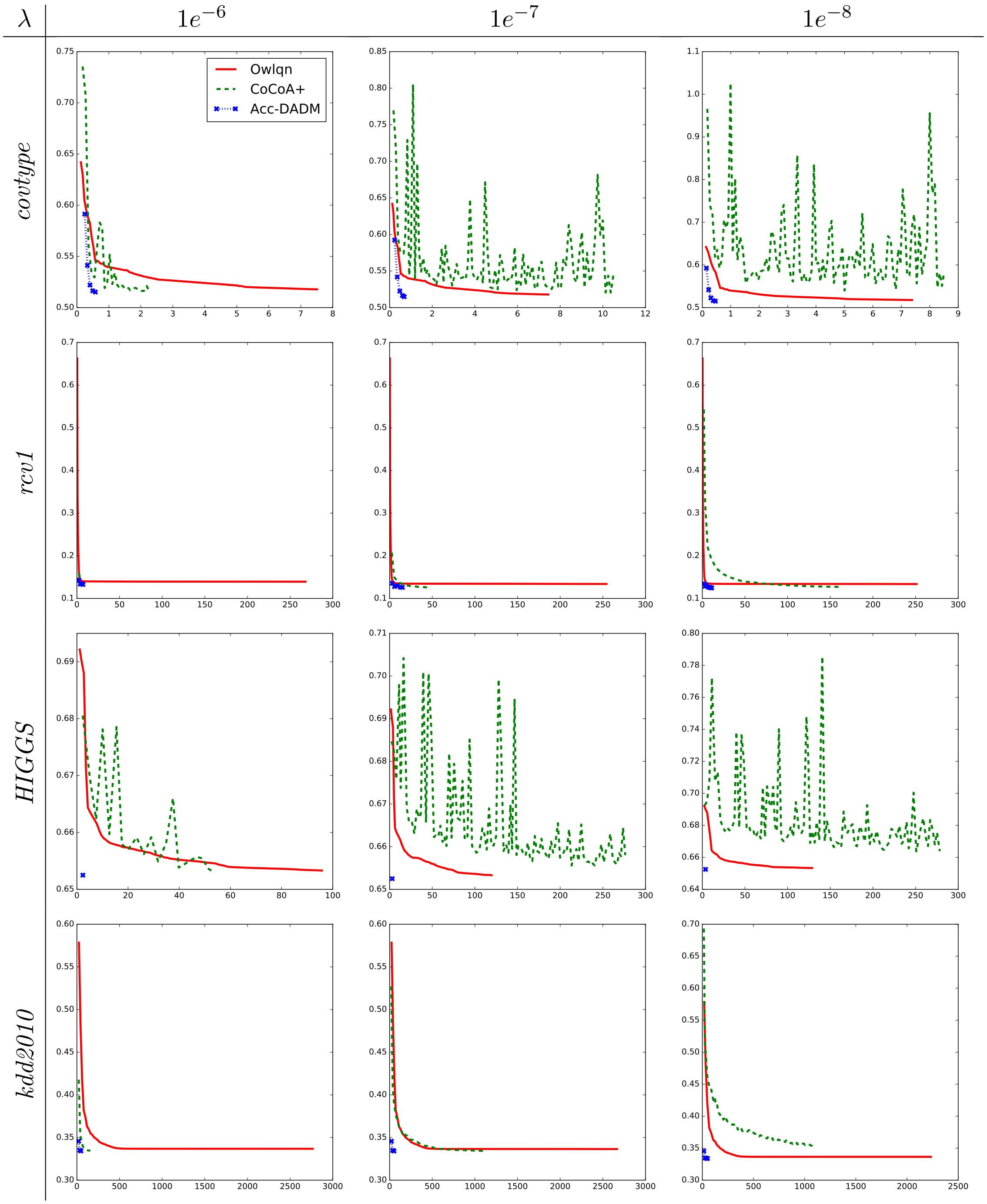

Figure 7: **The normalized primal objective versus time (s)** of **LR** experiments with $\mu = 1e^{-5}$, $\lambda$ varing in the range $\{1e^{-6}, 1e^{-7}, 1e^{-8}\}$ and $sp = 1.0$ on four datasets. We terminate methods either if the stopping condition is met ($1e^{-3}$ duality gap) or after 100 passes over the data.

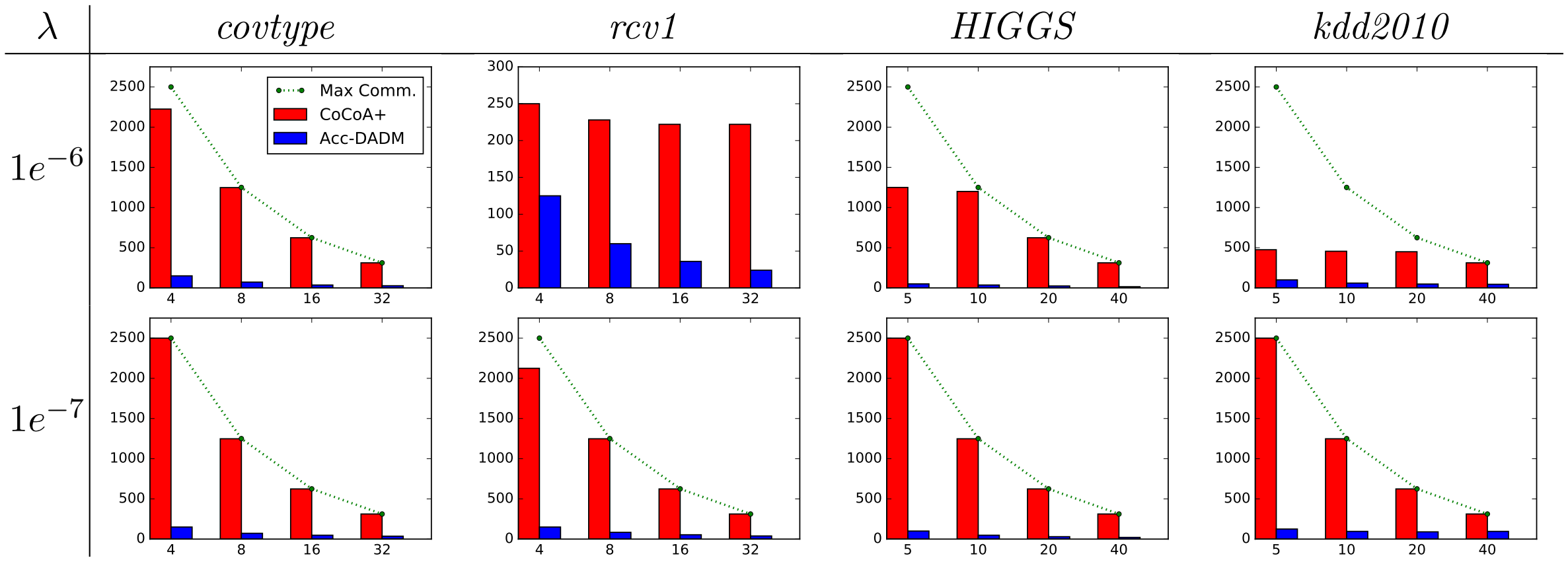

Figure 8: **The number of communications to reach $1e^{-3}$ duality gap versus the number of machines** of **SVM** experiments on four datasets. We fix the mini-batch size by varying *sp* as 0.04, 0.08, 0.16, 0.32 when $m$ grows exponentially from 4 to 32 or from 10 to 80. For each case, we run algorithms for at most 100 passes over the data. *Max Comm.* represents the total number of communications needed to go through 100 passes over the data.

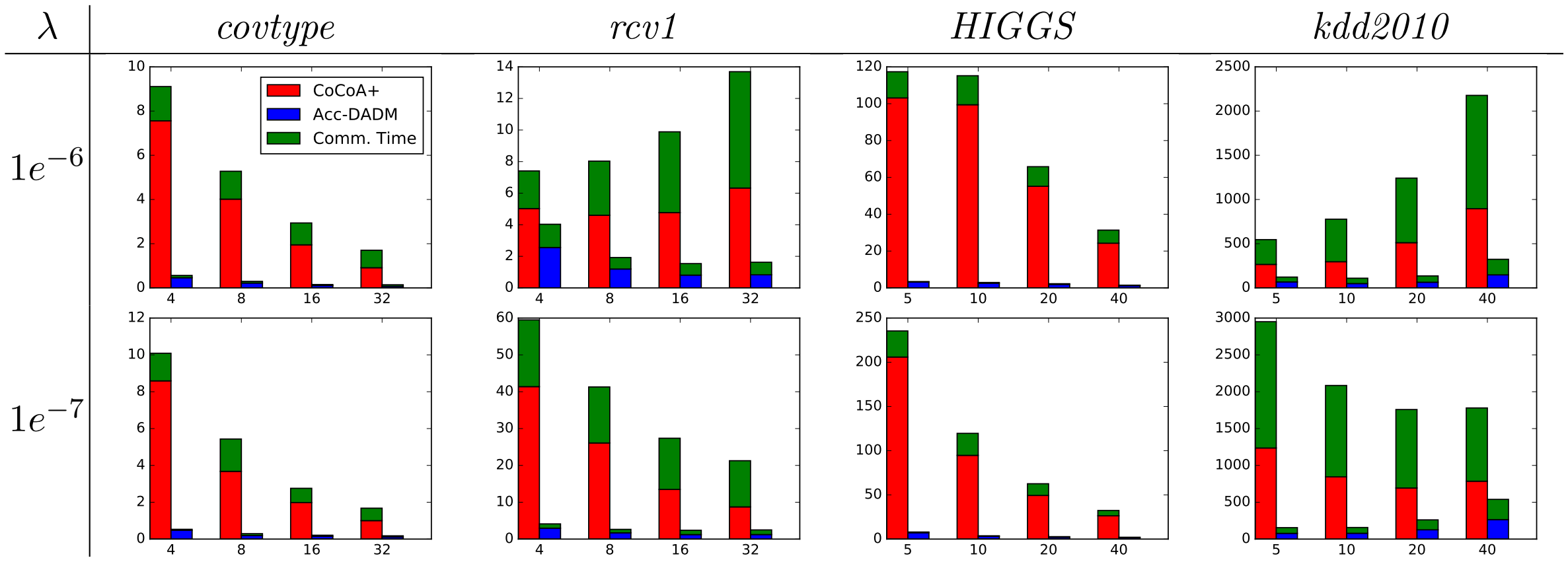

Figure 9: **Time (s) to reach $1e^{-3}$ duality gap versus the number of machines** of **SVM** experiments on four datasets. We fix the mini-batch size by varying *sp* as 0.04, 0.08, 0.16, 0.32 when $m$ grows exponentially from 4 to 32 or from 10 to 80. For each case, we run algorithms for at most 100 passes over the data. *Comm. Time* represents the total communication time of the corresponding algorithm.

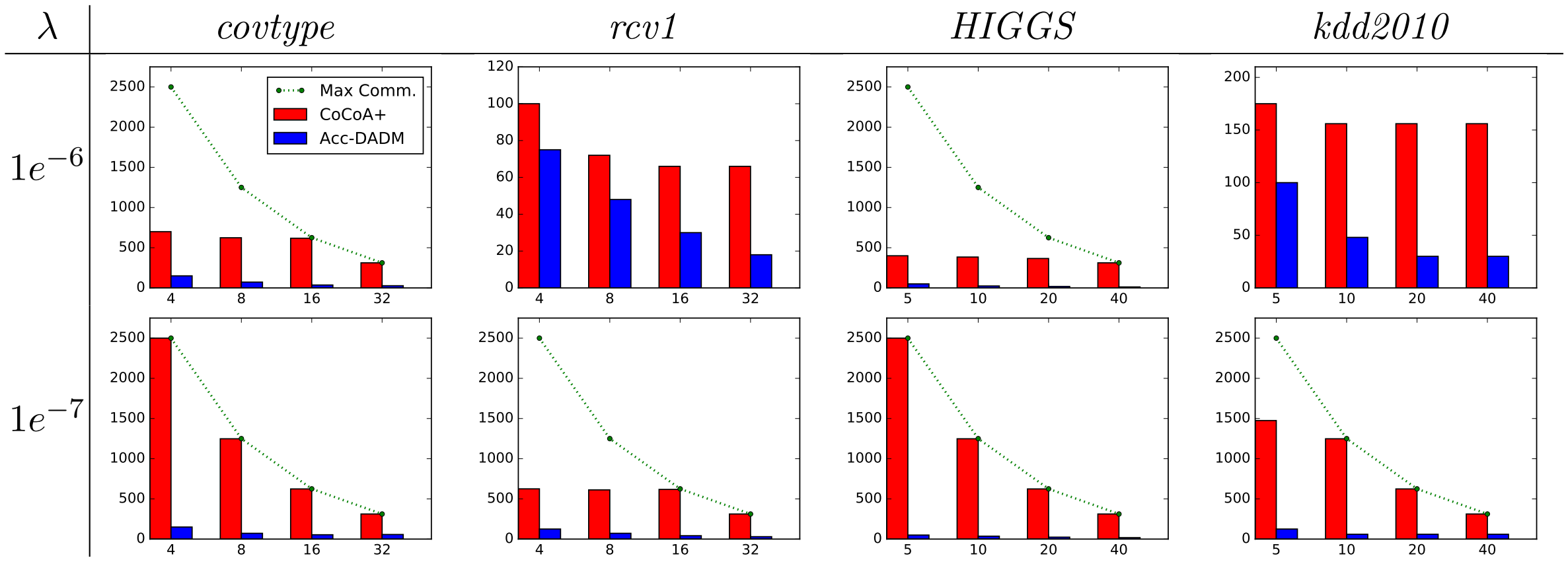

Figure 10: **The number of communications to reach $1e^{-3}$ duality gap versus the number of machines** of **LR** experiments on four datasets. We fix the mini-batch size by varying *sp* as 0.04, 0.08, 0.16, 0.32 when $m$ grows exponentially from 4 to 32 or from 10 to 80. For each case, we run algorithms for at most 100 passes over the data. *Max Comm.* represents the total number of communications needed to go through 100 passes over the data.

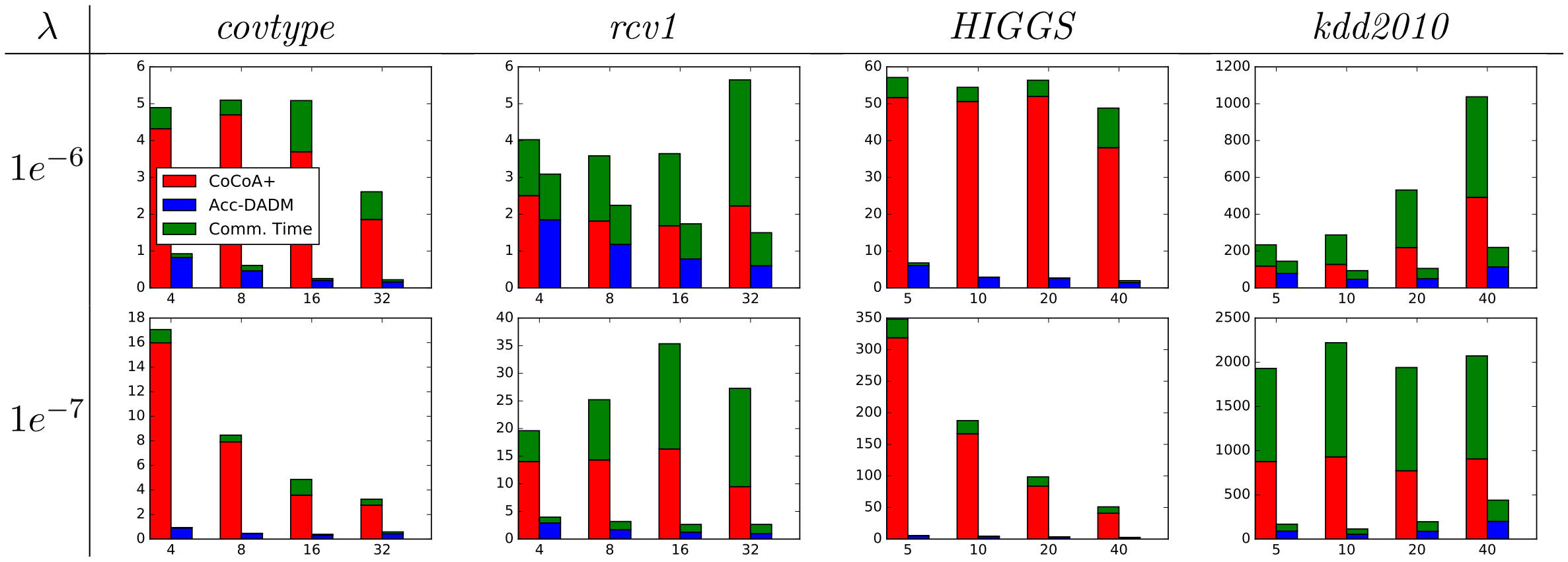

Figure 11: **Time (s) to reach $1e^{-3}$ duality gap versus the number of machines** of **LR** experiments on four datasets. We fix the mini-batch size by varying *sp* as 0.04, 0.08, 0.16, 0.32 when $m$ grows exponentially from 4 to 32 or from 10 to 80. For each case, we run algorithms for at most 100 passes over the data. *Comm. Time* represents the total communication time of the corresponding algorithm.

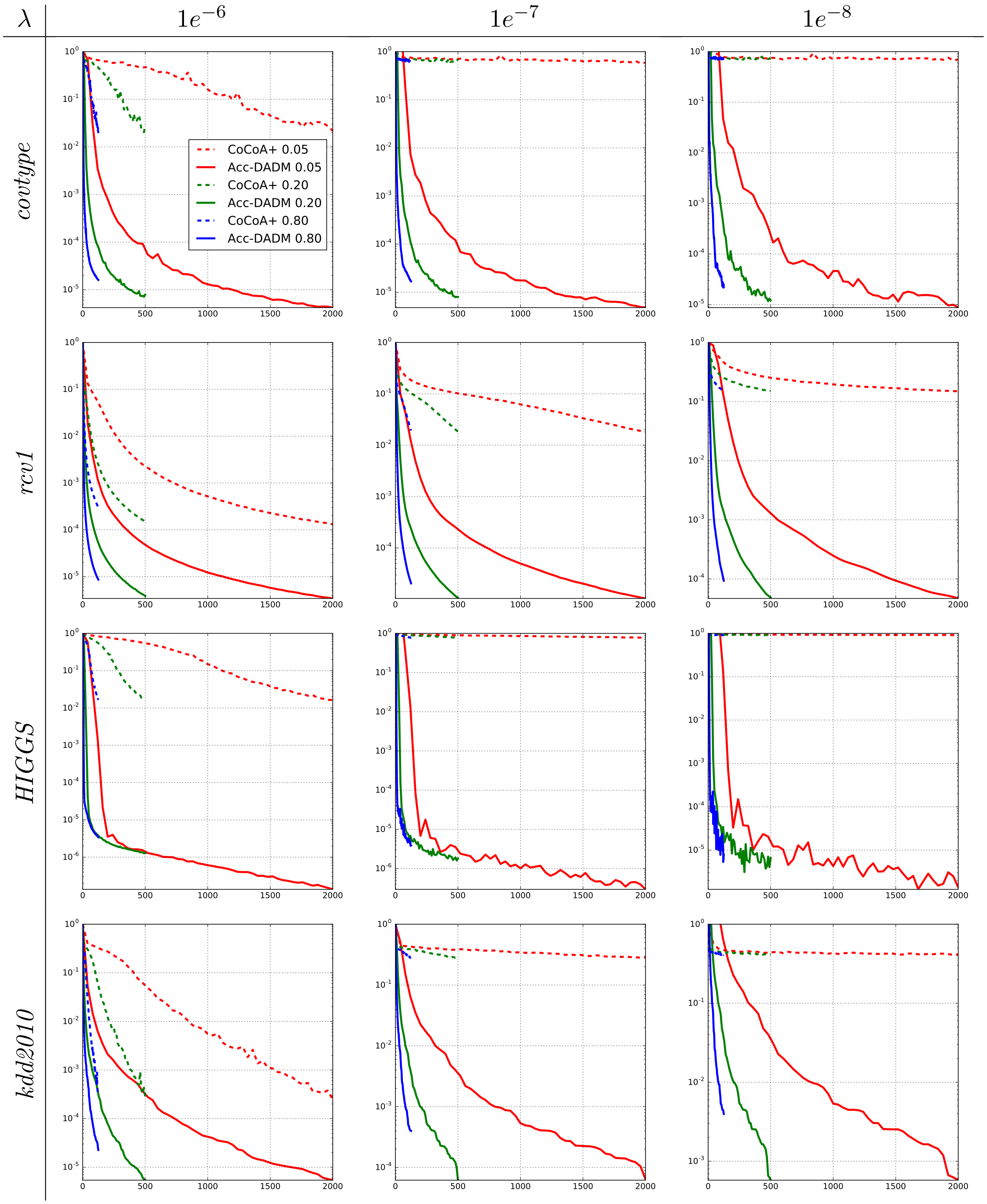

Figure 12: **The normalized duality gap versus the number of communications of the hinge loss** experiments with $\mu = 1e^{-5}$, $\lambda$ varing in the range $\{1e^{-6}, 1e^{-7}, 1e^{-8}\}$ and *sp* varing in the range $\{0.05, 0.20, 0.80\}$ on four datasets. We run methods in each case for 100 passes over the data.

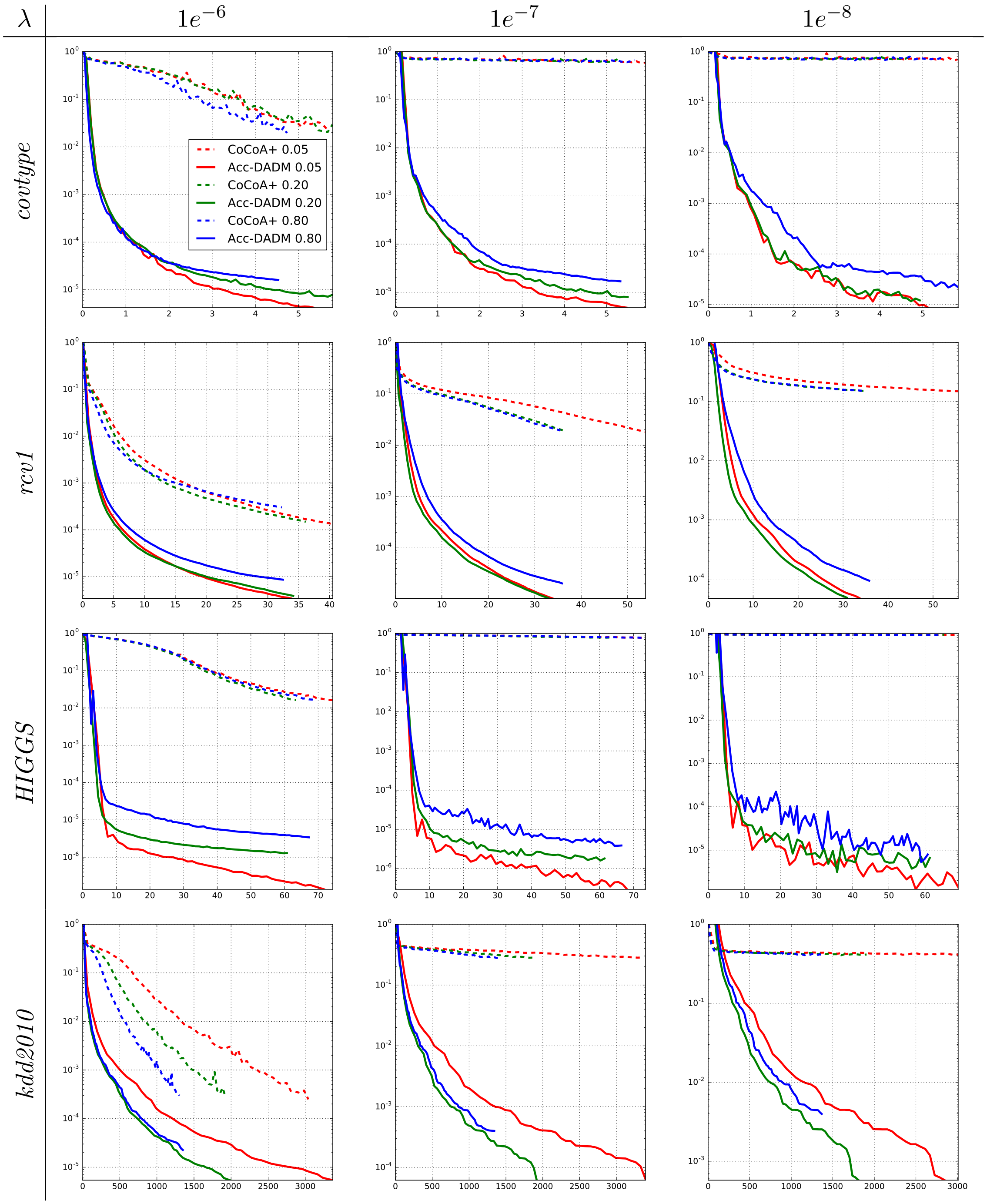

Figure 13: **The normalized duality gap versus time (s)** of **the hinge loss** experiments with $\mu = 1e^{-5}$, $\lambda$ varing in the range $\{1e^{-6}, 1e^{-7}, 1e^{-8}\}$ and $sp$ varing in the range $\{0.05, 0.20, 0.80\}$ on four datasets. We run methods in each case for 100 passes over the data.